\documentclass[times,review,10pt]{elsarticle}
\usepackage{amsfonts}

\usepackage{algorithm}
\usepackage{algorithmic}

\usepackage{xcolor}
\usepackage{soul}

\sethlcolor{yellow}

\soulregister\citep7
\soulregister\citet7
\soulregister\cite7
\soulregister\ref7
\soulregister\eqref7
\soulregister\emph7
\soulregister\textbf7

\usepackage{amsmath,amssymb,amsthm}
\usepackage{graphicx}
\usepackage{booktabs}      
\usepackage{array}
\usepackage{multirow}
\usepackage{makecell}
\usepackage{url}

\usepackage[
top=4.3cm,
bottom=4.3cm,
left=4.8cm,
right=4.8cm
]{geometry}

\usepackage{setspace}
\usepackage[font=footnotesize]{caption}

\usepackage{etoolbox}
\makeatletter
\patchcmd{\pprintMaketitle}
  {\normalsize\elsauthors\par\vskip10pt}
  {\footnotesize\elsauthors\par\vskip10pt}
  {}{}

\patchcmd{\MaketitleBox}
  {\normalsize\elsauthors\par\vskip10pt}
  {\footnotesize\elsauthors\par\vskip10pt}
  {}{}
\makeatother

\usepackage[percent]{overpic}
\usepackage{tikz}
\usepackage{adjustbox}
\usepackage{subcaption}

\usepackage{placeins}
\usepackage[hidelinks]{hyperref}
\usepackage{float}

\newcommand{\eqautoref}[1]{\hyperref[#1]{Eq.~(\ref*{#1})}}

\newcommand{\vx}{\mathbf{x}}

\newcommand{\vm}{\mathbf{m}}
\newcommand{\nm}{\textit{DeferredSeg}}
\newcommand{\gain}[1]{\textcolor{gray}{\,$\uparrow$#1}}

\newcommand{\drop}[1]{\textcolor{gray}{\,$\downarrow$#1}}

\journal{Pattern Recognition}

\begin{document}

\begin{frontmatter}

\begin{graphicalabstract}
  \centering
  \includegraphics[width=\textwidth]{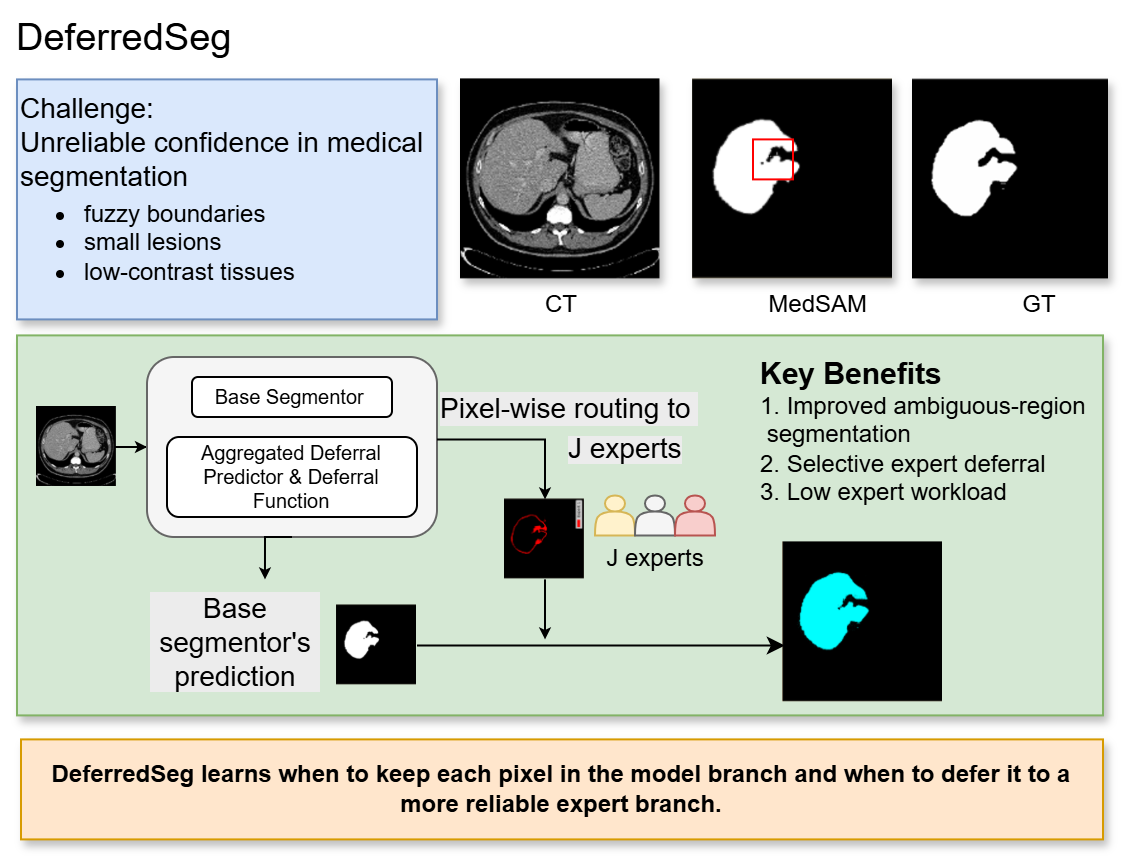}
\end{graphicalabstract}


\begin{highlights}
\item $\nm$, a pixel-wise deferral framework for medical image segmentation.
\item A Deferral Collaboration loss learns pixel-wise routing between the model and expert branches.
\item An aggregated deferral predictor encourages spatially coherent deferral regions.
\item A load-balancing penalty mitigates routing collapse in multi-expert settings.
\item Experiments with synthetic expert masks and offline clinician annotations evaluate  collaborative segmentation.
\end{highlights}

\title{$\nm$: A Multi-Expert Deferral Framework for Medical Image Segmentation}

\tnotetext[t1]{The work is supported in part by the Natural Science Foundation of China (No.~U23A20389), the Natural Science Foundation of Shandong Province (No.~ZR2024MF101), the Young Expert of Taishan Scholars (No.~tsqn202312026), and Shandong Sci-tech SMEs Innovation Project (No.~2024TSGC0740).}

\author[inst1]{Qiuyu Tian}
\ead{202435328@mail.sdu.edu.cn}

\author[inst1]{Haoliang Sun\corref{cor1}}
\ead{haolsun@sdu.edu.cn}

\author[inst2]{Yunshan Wang}
\ead{wangyunshan135@126.com}

\author[inst3]{Yinghuan Shi}
\ead{syh@nju.edu.cn}

\author[inst1]{Yilong Yin}
\ead{ylyin@sdu.edu.cn}

\cortext[cor1]{Corresponding author.}

\affiliation[inst1]{
  organization={School of Software, Shandong University},
  city={Jinan},
  postcode={250101},
  country={China}
}

\affiliation[inst2]{
  organization={Department of Clinical Laboratory, Shandong Provincial Hospital Affiliated to Shandong First Medical University},
  city={Jinan},
  postcode={250021},
  country={China}
}

\affiliation[inst3]{
  organization={School of Computer Science and Technology, Nanjing University},
  city={Nanjing},
  postcode={210023},
  country={China}
}

\begin{abstract}

Segmentation models based on deep neural networks demonstrate strong generalization for medical image segmentation. However, they often exhibit overconfidence or underconfidence, leading to unreliable confidence scores for segmentation masks, especially in ambiguous regions. This limits the reliability of fully automatic segmentation in ambiguous regions. Motivated by the learning-to-defer (L2D) paradigm, we introduce $\nm$,
a pixel-wise Human--AI collaborative deferral framework that learns whether to retain model predictions or defer difficult pixels or regions to expert branches.

$\nm$ extends the base segmentor with an aggregated deferral predictor and additional routing channels that dynamically route each pixel to either the base segmentor or human expert. To train this routing efficiently, we introduce a pixel-wise surrogate collaboration loss that supervises deferral decisions. In addition, we introduce a spatial-coherence loss that aligns the aggregated expert-assignment probability with the learned routing decisions, encouraging locally consistent deferral patterns.

Beyond single-expert deferral, we further extend the framework to a multi-expert setting by introducing multiple experts with different reliability profiles for collaborative decision-making. To mitigate overloading or underutilization of individual experts, we further design a load-balancing penalty that discourages routing collapse and promotes more balanced utilization across expert branches. We evaluate $\nm$ on multiple medical segmentation benchmarks using MedSAM and CENet as the base segmentors, and further compare it with an expert-assisted patch-correction baseline. Experimental results show that $\nm$ consistently improves System-level segmentation performance while selectively deferring difficult pixels or regions to more reliable expert branches. Moreover, the proposed framework is backbone-compatible and can be instantiated with different segmentation architectures that provide intermediate feature representations.
\end{abstract}


\end{frontmatter}

\section{Introduction}
\label{sec:introduction}

Medical image segmentation is a fundamental task in medical data analysis, aiming to assign a semantic label to every pixel or voxel in an image.  
Accurate delineation of anatomical structures or pathological regions is crucial for a broad range of downstream applications, including computer-aided diagnosis, treatment planning, and intraoperative navigation~\citep{litjens2017survey}. Compared to natural image segmentation, medical segmentation is inherently more challenging due to heterogeneous imaging modalities, low-contrast boundaries, irregular shapes, and the scarcity of high-quality annotations.
Traditional approaches have demonstrated effectiveness, but deep neural networks now dominate medical image segmentation. Skip-connection architectures, notably U-Net~\citep{ronneberger2015unet} and its variants~\citep{oktay2018attention,kamnitsas2017efficient}, remain strong clinical baselines. Recent Transformer-based designs, including kMaXU~\citep{huang2025kmaxu} , further enhance performance through global-context and multi-scale representation learning.

Although these models have demonstrated remarkable generalization ability, recent studies report that these deep models often produce \emph{overconfident} or \emph{underconfident} predictions~\citep{lin2025uncertainty}. This tendency is particularly evident in ambiguous or low-contrast regions, such as fuzzy boundaries, small lesions, and distribution shifts~\citep{mehrtash2020confidence}. In high-stakes medical segmentation, such issues pose a critical risk: incorrect predictions delivered with high certainty may mislead clinicians and cause severe consequences.

Therefore, relying solely on automated outputs is unsafe; incorporating \emph{Human--AI collaboration} that integrates clinical expertise into the decision-making process for high-risk cases is essential. Several interactive segmentation methods have been proposed, such as MIDeepSeg~\citep{luo2021mideepseg} and TIS~\citep{liu2022transforming}, which reduce the amount of user interaction required. However, these methods still require the user to make decisions on the entire image, rather than focusing on smaller, more ambiguous regions.

The learning-to-defer paradigm addresses this issue by allowing a model to either make its own prediction or defer to an expert when confidence is low~\citep{mozannar2020consistent,wei2024exploiting}. Unlike conventional uncertainty estimation, L2D treats referral as an
explicit decision and routes unreliable predictions to experts
~\citep{de2021classification}.

However, existing L2D methods operate at the instance level, making them unsuitable for dense prediction tasks where uncertainty varies spatially at the pixel level. In medical segmentation, such instance-wise deferral cannot resolve localized ambiguities, including unclear tissue boundaries and heterogeneous lesion textures. This motivates a region-selective collaboration strategy for dense prediction, where only ambiguous regions are referred for expert review instead of requiring full-image correction.

\begin{figure}[t]
    \centering
    \includegraphics[width=\linewidth, height=0.18\textheight, keepaspectratio]{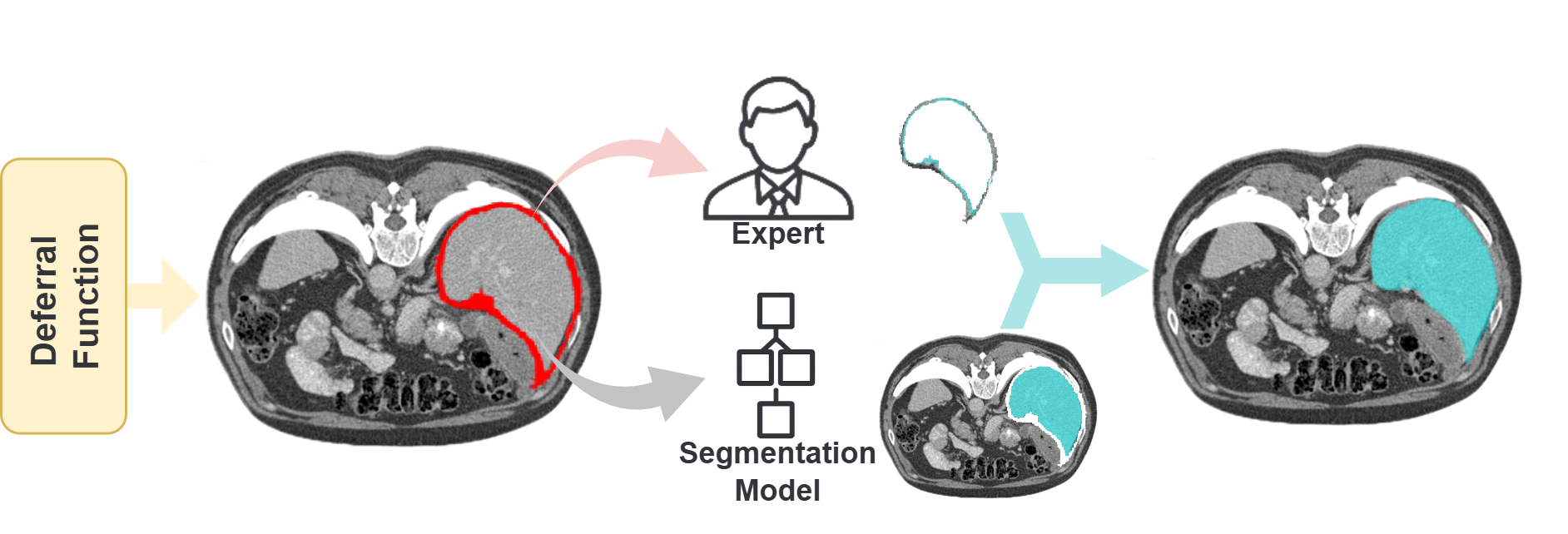}
    \caption{Overview of $\nm$. A deferral predictor produces a pixel-wise deferral map that triages image regions between human-expert branches and the base segmentation model, enabling spatially adaptive Human--AI collaboration for deferral.}
    \label{fig:intro}
\end{figure}

In this paper, we propose $\nm$, a pixel-wise learning-to-defer framework for dense medical image segmentation.
 The core of our framework is a pixel-wise deferral mechanism designed to support Human--AI collaboration through a deferral predictor and a learned routing function. These components dynamically route pixel-wise predictions, determining whether to accept the base segmentation model's prediction or to defer ambiguous regions to human experts, as illustrated in Fig.~\ref{fig:intro}. In the intended deployment, experts provide local corrections only for the
pixels or regions selected by the learned deferral function.

Our main contributions are:
\begin{enumerate}
    \item \textbf{A Novel Deferral-Aware Segmentation Framework.}
    We introduce $\nm$, a pixel-wise deferral-aware framework for Human--AI collaborative medical image segmentation. It features a dynamic deferral predictor that intelligently routes pixel-wise decisions to either the base model or expert branches. This is enabled by two tailored loss functions: a surrogate deferral collaboration loss for efficient training and a spatial-coherence loss to ensure deferred regions are spatially coherent and reliable.
    \item \textbf{A Multi-Expert Setting for Balanced Collaboration.}
    Our method introduces multiple experts to facilitate collaborative decision-making. This includes a load-balancing penalty that mitigates routing collapse and discourages severe overload or underutilization of individual expert branches.
    \item \textbf{Backbone-compatible Instantiation.}
We instantiate $\nm$ with MedSAM and further validate its transferability on CENet, showing that the proposed routing mechanism can be attached to different segmentation backbones.
    \item \textbf{Comprehensive Evaluation.}
Experiments with synthetic expert masks and pre-existing clinician
annotations show that $\nm$ improves System-level segmentation performance
while quantifying pixel-level expert-routing ratios.
\end{enumerate}

\section{Related Work}
\label{sec:related_work}
\subsection{Medical Image Segmentation}

\subsubsection{Classical and Deep Learning Methods}
Early medical image segmentation relied on handcrafted methods such as energy-based models, threshold-based methods~\citep{sezgin2004survey}, and graph-based methods~\citep{shi2000normalized}. Although effective for high-contrast structures, these methods are sensitive to noise, parameter tuning, and anatomical variability, and have gradually been replaced by data-driven learning approaches~\citep{huang2024medical}.

Building on the limitations of classical methods, learning-based approaches, especially CNN-based architectures, have become dominant in medical image segmentation. Representative models include U-Net~\citep{ronneberger2015unet}, DeepMedic~\citep{kamnitsas2017efficient}, and the self-configuring nnU-Net framework~\citep{isensee2021nnu}, while variants such as Attention U-Net~\citep{oktay2018attention} improved boundary delineation. More recently, Transformer-based models such as UNETR~\citep{hatamizadeh2022unetr} further improve long-range dependency modeling. Hybrid CNN--Transformer architectures have also been developed to combine local texture modeling with global context modeling, such as CTC-Net~\citep{yuan2023ctcnet} and TBConvL-Net~\citep{iqbal2025tbconvl}. Despite these advances, robust segmentation remains challenging for low-contrast tissues, irregular boundaries, and rare anatomical structures~\citep{litjens2017survey}.

\subsubsection{Foundation Models for Image Segmentation}
Foundation models have recently attracted increasing attention in image segmentation due to their strong generalization capabilities and potential for zero-shot or few-shot transfer across diverse modalities and anatomical structures. Among them, the Segment Anything Model (SAM)~\citep{kirillov2023segment} enables flexible prompt-based mask generation using spatial prompts such as points, bounding boxes, and prior masks. Inspired by SAM's success in natural image segmentation, several works have adapted this paradigm to the medical domain. MedSAM~\citep{ma2024segment} fine-tunes SAM on a large corpus of medical images while preserving its prompt-based architecture, establishing a strong baseline for generalizable medical segmentation. MA-SAM~\citep{yan2022sam} improves spine segmentation through
multi-atlas-guided pseudo-mask prompts, while
SegMIC~\citep{zhao2025segmic} uses in-context learning to support
generalizable medical image segmentation. Despite these advances, prompt-based foundation models still face challenges in clinically critical scenarios involving low-contrast regions, irregular boundaries, small lesions, and ambiguous structures.

\subsubsection{Human--AI-assisted and Multi-expert Segmentation}

Human--AI-assisted segmentation improves reliability by incorporating user interaction, expert review, or annotation feedback into the segmentation process. Interactive segmentation methods refine masks using clicks, scribbles, bounding boxes, or other prompts. Representative methods such as MIDeepSeg~\citep{luo2021mideepseg} and TIS~\citep{liu2022transforming} show that explicit interaction can improve segmentation quality, but they usually require users to inspect the image and provide interaction signals during inference, which may still be time-consuming for scattered errors or ambiguous structures.

Uncertainty-based referral limits expert review to unreliable or
informative samples or regions. Uncertainty-aware segmentation architectures, such as UCTNet~\citep{guo2024uctnet}, further exploit uncertainty to guide feature extraction or model computation in medical image segmentation. Uncertainty-based methods often rely on confidence scores, entropy, prediction variance, or model disagreement to identify unreliable predictions~\citep{hasan2025survey}. 

Patch-based error correction provides a local form of expert intervention. For example, PatchCorr~\citep{koehler2024efficiently} selects uncertain patches and corrects them with expert input, avoiding full-image manual correction. However, such methods usually operate at a coarse patch level and depend on uncertainty-guided selection, without explicitly learning pixel-wise routing among the model and multiple expert branches. Moreover, the correction ratio or expert-intervention budget must be manually specified in advance, which may require task-specific tuning.

Expert specialization has also been explored in MoE-style medical segmentation. Recent methods such as MoE-SAM~\citep{li2025moesam},
topology-preserving MoE segmentation~\citep{ma2026topology}, and medical
language MoE~\citep{liu2024medical} demonstrate the value of specialized
computational experts in medical image analysis. These methods demonstrate the value of specialized experts, but their experts are typically computational modules inside the network rather than external expert branches used for inference-time deferral. 

Overall, existing studies have advanced interactive correction, uncertainty-guided review, active annotation, patch-level quality control, and computational expert specialization. However, most of them rely on predefined interaction or uncertainty-selection strategies, or use experts as internal model modules. Therefore, they do not directly address how to automatically decide, at the pixel level, which regions should remain with the segmentation model and which difficult regions should be assigned to external human expert branches. This gap motivates a pixel-wise deferral formulation for medical image segmentation.

\subsection{Learning to Defer}
\label{sec:l2d}
\textit{L2D} integrates algorithmic or human expertise into automated decision systems by learning whether to make a prediction or defer the input to one or more experts. It extends the classic \emph{learning to reject} paradigm~\citep{bartlett2008classification} by replacing simple rejection with expert-aware triage. Madras et al.~\citep{madras2018predict} formulated learning to defer to human experts, while De et al.~\citep{de2020regression} further extended deferral to regression.

Direct optimization of the discontinuous deferral objective is intractable, so L2D research has focused on designing consistent, differentiable surrogates. A cost-sensitive, consistent surrogate loss using a softmax-based formulation was first proposed in \cite{mozannar2020consistent}. A calibrated one-vs-all objective was introduced in \cite{verma2022calibrated}. For multi-expert scenarios, a consistent and calibrated surrogate was developed in \cite{verma2023learning}, and a two-stage approach with a matching consistent loss was proposed in \cite{mao2023two}.

L2D has also been extended beyond classification, including regression with expert deferral~\citep{mao2024regression}, sequential healthcare decision-making~\citep{strong2024towards} and semi-supervised learning with sparse expert availability~\citep{nguyen2025probabilistic}. These studies demonstrate the generality of L2D for Human--AI collaboration across tasks.

However, most L2D methods make instance-level routing decisions, which are unsuitable for dense segmentation where uncertainty varies spatially and errors concentrate along fine anatomical boundaries (e.g., fuzzy borders, low-contrast tissues, small lesions). This motivates our pixel-wise deferral formulation for medical image segmentation.

\section{Preliminaries}
\label{sec:preliminaries}
We begin by formalizing the \textit{general} learning-to-defer problem with multiple experts, and then provide the formulation of the true risk for the 0--1 loss and its surrogates for optimization.

\subsection{Data and Model}
Suppose there are $J$ experts involved in L2D tasks. For a $K$-class classification problem, each training example in L2D is represented as a triplet $(\vx, y, \vm) \sim \mathcal{D}$, where $\vx \in \mathcal{X}$ is the input, $y \in \{1, \dots, K\}$ is the ground-truth label, and $\vm \in \mathbb{R}^J$ is the vector of expert's prediction with element $\vm_i \in \{1, \dots, K\}$. In addition to a classifier $h: \mathcal{X} \rightarrow \mathbb{R}$, a deferral function $r: \mathcal{X} \rightarrow \{0,1,\dots,J\}$ determines whether to defer and, if so, to which expert. Specifically, when $r(\vx) = 0$, the classifier makes the prediction; when $r(\vx) = j$, the classifier abstains and defers the decision to the $j$-th expert.

\subsection{The True Risk of the 0--1 Loss}
The true risk \( \mathcal{L} \) of L2D is extended from a generalization of learning with rejection, as proposed in \cite{cortes2016learning}, which can be formalized as a 0--1 loss as follows:
\begin{equation}
\mathcal{L}_{0-1}(h,r)=
\mathop{\mathbb{E}}\limits_{\vx,y,\vm}\!\Big[
  \mathop{\mathbb{I}}\limits_{r(\vx)=0} ~\mathop{\mathbb{I}}\limits_{h(\vx)\neq y}
  \;+\;
  \sum_{j=1}^{J}
  \mathop{\mathbb{I}}\limits_{r(\vx)=j} ~\mathop{\mathbb{I}}\limits_{\vm_j\neq y}
\Big].
\label{0-1_loss}
\end{equation}

Here, \(\mathbb{I}[\cdot]\) denotes the indicator function, which equals 1 if the condition inside the brackets is satisfied and 0 otherwise.

The estimated classifier and deferral function pair \( (\hat{h}, \hat{r}) \) can be obtained by minimizing Eq. (\ref{0-1_loss}) on the training set \( \mathcal{D} \). However, $\mathcal{L}_{0\text{-}1}$ is discontinuous and non-convex, making direct minimization impractical; therefore, we  adopt differentiable and consistent surrogates for optimization.

\subsection{The Softmax-Based (SM) Surrogate Loss}
The softmax-based loss is commonly used in L2D. We augment the label space as \( \mathcal{Y}^\perp = \mathcal{Y} \cup \{ \perp_1, \dots, \perp_J \} \), where \( \perp_j \) represents the option to defer to the \( j \)-th expert. 
We build a predictor \( f: \mathcal{X} \rightarrow \mathbb{R}^{K+J} \), composed of the classifier \( h \) and deferral function \( r \). \( f \) can also be decomposed into \( K + J \) functions: \( f_k: \mathcal{X} \rightarrow \mathbb{R} \) for \( k\in\{1,\dots,K\} \), and \( f_{\perp,j}: \mathcal{X} \rightarrow \mathbb{R} \) for \( j\in\{1,\dots,J\} \), where \( k, j \) denote the class and expert indexes, respectively. These \( K + J \) functions are combined using a softmax-parameterized surrogate loss:
\begin{align}\label{eq:sm_loss} 
    \mathcal{L}_{SM}^J&(f_1, \dots, f_K, f_{\perp,1}, \dots, f_{\perp,J}; \vx, y, \vm) =  \nonumber \\ 
    &- \sum_{j=1}^{J} \mathbb{I} [ \mathbf{m}_j = y ] \log \left( \frac{\exp f_{\perp,j}(\vx)}{\sum_{y' \in \mathcal{Y}^\perp} \exp f_{y'}(\vx)} \right)
    - \log \left( \frac{\exp f_y(\vx)}{\sum_{y' \in \mathcal{Y}^\perp} \exp f_{y'}(\vx)} \right).
\end{align}
The first term maximizes the deferral function \( f_{\perp,j} \) if the \( j \)-th expert's prediction is correct. The second term ensures that the function \( f_k \) is maximized for the true label. At test time the classifier prediction is obtained by selecting the largest score among the first $K$ outputs, and deferral is decided as follows:

\begin{equation}
r(\vx)=
\begin{cases}
0, & \text{if } f_h(\vx)  > \max_{j\in\{1,\dots,J\}} f_{\perp,j}(\vx),\\
\arg\max_{j\in\{1,\dots,J\}} f_{\perp,j}(\vx), & \text{otherwise.}
\end{cases}
\label{eq:routing_function}
\end{equation}

\section{Method}\label{sec:method}
Our methodology extends learning to defer (L2D) from instance-level prediction to pixel-wise medical image segmentation. This section presents the overall framework of $\nm$, the pixel-wise routing formulation, the training objectives, and the MedSAM-based instantiation.

\subsection{Overall Framework}\label{sec:framework}
$\nm$ is a backbone-compatible pixel-wise deferral framework with multi-expert collaboration, designed to work with a base segmentation model that provides dense feature representations and segmentation predictions. We augment the model's output channels by adding multiple routing channels, where one corresponds to the base model and the remaining correspond to specialized expert branches. These routing channels provide the architectural basis for pixel-wise deferral by producing branch-specific routing logits across spatial regions. 
An auxiliary CNN-based aggregated deferral predictor enforces spatially
coherent routing, and training combines collaboration, spatial-coherence,
and load-balancing losses.

Given an input image, the base segmentor first produces dense features and segmentation predictions. The routing function then assigns each pixel either to the model branch or to an expert branch. The final System output retains the base prediction for model-routed pixels and uses the selected expert prediction for deferred pixels.
\subsection{Pixel-Wise L2D with Multiple Experts}
\subsubsection{Output Channel Extension and Pixel-wise Routing}\label{sec:SingleRouting}

Let $\vx\in\mathcal{X}$ be an image of size $H\times W$, and let $i\in\{1,\dots,HW\}$ index pixels; $\vx^i$ denotes the value at the $i$-th pixel. The ground-truth label of $\vx^i$ is $y^{i}\in\{0,1\}$ for a binary segmentation task. To enable pixel-wise L2D with $J$ experts, we extend the base segmentor by adding $J{+}1$ routing channels. The original segmentation output channels remain unchanged. 
Therefore, the label space is defined 
\(\mathcal{Y}^\perp = \mathcal{Y} \cup \{\perp_0, \perp_1, \cdots, \perp_J\}\), 
where the labels from \(\perp_0\) to \(\perp_J\) correspond to the additional routing channels. Unlike the classification setting in Sec.~\ref{sec:preliminaries}, in our segmentation setting, the predictor is composed of the base segmentor $f_{\text{seg}}:\mathcal{X}\to\mathbb{R}^{H\times W}$ and the deferral function $f_{\perp}:\mathcal{X}\to\mathbb{R}^{(J+1)\times H\times W}$. In practice, the base segmentor produces a probability prediction $\hat{y}^{i}\in[0,1]$ for the foreground class at each pixel $\vx^i$.
The deferral function outputs per-pixel logits $f_{\perp,j}(\vx^{i})$ with $j\in\{0,\dots,J\}$. A larger $f_{\perp,0}(\vx^{i})$ indicates a stronger preference to route pixel $i$ to channel $0$ (the base segmentor), 
while a larger $f_{\perp,j}(\vx^{i})$ ($j>0$) indicates a stronger preference to route it to channel $j$ (the $j$-th expert).
For the $j$-th expert branch, $\mathbf{m}_j^{i} \in \{0,1\}$ denotes the $j$-th expert’s binarized prediction at pixel $\vx^i$.

\subsubsection{The Deferral Collaboration Loss}\label{sec:loss}

To enable pixel-wise L2D, we adapt the softmax surrogate introduced in Sec.~\ref{sec:preliminaries} (Eq.~\eqref{eq:sm_loss}) and generalize it as the \emph{Deferral Collaboration (DC) Loss} for dense segmentation, which is defined as:
\begin{align}\label{eq:softmax_loss_multi} 
    \mathcal{L}_{\text{DC}}&(f_{seg}, f_{\perp,0}, \dots, f_{\perp,J}; 
    \vx^i, y^i, \vm^i)
    =\frac{1}{HW}\sum_{i=1}^{HW} \Bigg[ \nonumber
    -\sum_{j=1}^{J}
    \mathbb{I} [ \mathbf{m}_j^{i} = y^{i} ]
    \log \left(
    \frac{\exp f_{\perp,j}(\vx^i)}
    {\sum_{t=0}^{J} \exp f_{\perp,t}(\vx^i)}
    \right) \nonumber \\
    &-\sum_{j=1}^{J}
    \left(1-\mathbb{I} [ \mathbf{m}_j^{i} = y^{i} ]\right)
    \log \left(
    1-
    \frac{\exp f_{\perp,j}(\vx^i)}
    {\sum_{t=0}^{J} \exp f_{\perp,t}(\vx^i)}
    \right) \nonumber \\
    &-\mathbb{I}[\tilde{y}^{i} = y^{i}]
    \log \left(
    \frac{\exp f_{\perp,0}(\vx^i)}
    {\sum_{t=0}^{J} \exp f_{\perp,t}(\vx^i)}
    \right)
    -\Big(
    y^{i}\log \hat{y}^{i}
    +\big(1-y^{i}\big)\log\big(1-\hat{y}^{i}\big)
    \Big)\Bigg].
\end{align}
where \(\mathbb{I}[\cdot]\) denotes the indicator function, $\tilde{y}^{i}=\mathbb{I}[\hat{y}^{i}\ge 0.5]$ denotes binarized segmentor prediction. Specifically:
\begin{itemize}
    \item The first term increases the routing probability assigned to expert \(j\) when its prediction matches the ground truth. The second term suppresses the routing probability assigned to expert \(j\) when its prediction is incorrect.
     \item The third term maximizes the segmentor branch deferral score \(f_{\perp,0}(\vx^{i})\) if the segmentor prediction \(\tilde{y}^{i}\) matches the ground truth.
    \item The fourth term is the pixel-wise binary cross-entropy (BCE) on the segmentor output \(\hat{y}^{i}\), which supervises the foreground prediction itself.
\end{itemize}

The loss encourages each pixel to select the branch with the more reliable prediction. The thresholded prediction \(\tilde{y}^{i}\) is used only to construct routing supervision, whereas the segmentation branch itself is optimized through the soft probability \(\hat{y}^{i}\) in the BCE term.

During inference, the routing probability map $\mathcal{M}$ of the $j$-th expert at pixel \(i\) is obtained by applying a softmax over all routing logits:
\begin{equation}
\mathcal{M}_j^{\,i}
\;=\;
\frac{\exp\!\big(f_{\perp,j}(\vx^{i})\big)}{\sum_{j=0}^{J}\exp\!\big(f_{\perp,j}(\vx^{i})\big)}.
\label{eq:alpha_j_pixel}
\end{equation}

\subsubsection{Spatial-Coherence Loss}\label{sec:LG_1}
To generate a smooth deferral map and ensure that deferred regions are spatially coherent and reliable, we introduce an additional \emph{aggregated deferral predictor} built with two convolutional layers and enforce consistency between its predictions and those of the deferral function score $f_{\perp,j}(\cdot)$ using a \emph{spatial coherence (SC) loss}.

\textbf{Motivation:}\emph{
State-of-the-art medical image segmentation models often employ Vision Transformer (ViT)-based decoders. However, these architectures can be less effective for data-efficient deferral learning on small datasets, as ViTs exhibit weak inductive bias for spatial relevance due to their emphasis on long-range interactions~\citep{lu2022bridging}. To address this, we introduce a CNN-based predictor to complement the ViT by providing intra-patch locality, enabling the capture of fine structures within patches and improving the spatial consistency of expert-assignment regions.}

Let \(\tilde{\mathcal{M}} \in [0,1]^{H\times W}\) denote the aggregated expert-assignment probability map predicted by the aggregated deferral predictor.
 Given the per-pixel routing probabilities $\mathcal{M}_j^{\,i}$, each pixel of the binary supervision mask $\bar{\mathcal{M}}$ is computed as 
\begin{equation}
\bar{\mathcal{M}}^{\,i}\;=\;\begin{cases}
  0, &  \text{if}  \;\; \operatorname*{arg\,max}_{j\in\{0,\dots,J\}}  \mathcal{M}_j^{\,i} = 0  \\
  1,  & \text{otherwise.}
\end{cases} 
\label{eq:onehot_expert_pixel}
\end{equation}

Here, \(\bar{\mathcal{M}}\) is not a ground-truth segmentation mask; instead, it serves as a pseudo supervision signal indicating whether a pixel is routed to the model branch or to any expert branch according to the current routing logits. The role of the SC loss is therefore not to recover semantic labels, but to encourage spatial consistency in the deferral pattern. The goal of the \emph{spatial coherence (SC) loss} is to minimize the discrepancy between $\tilde{\mathcal{M}}$ and $\bar{\mathcal{M}}$:
\begin{equation}
\begin{aligned}
\mathcal{L}_{\text{SC}}
=  \frac{1}{HW}\sum_{i=1}^{HW} \Big[\, \underbrace{-\left(\bar{\mathcal{M}}^{\,i}\log \tilde{\mathcal{M}}^{\,i} + (1-\bar{\mathcal{M}}^{\,i}) \log (1-\tilde{\mathcal{M}}^{\,i})\right)}_{\text{BCE}} 
 + \beta_1 \underbrace{\left( \tilde{\mathcal{M}}^{\,i} - \sum_{j=1}^{J}\mathcal{M}_{j}^{\,i} \right)^{2} }_{\text{MSE}}
\; + \; \beta_2\,\tilde{\mathcal{M}}^{\,i}\Big].
\end{aligned}
\label{eq:coherence_loss}
\end{equation}

We adopt two commonly used losses with different properties. The first term is the pixel-wise binary cross-entropy (BCE). The second term is the mean squared error (MSE) aligning \(\tilde{\mathcal{M}}^{\,i}\) with the sum of all expert routing probabilities \(\mathcal{M}_{j}^{\,i}\). 
Additionally, the last term penalizes excessive deferral and discourages the trivial all-defer solution. $\beta_1$ and $\beta_2$ are two weighting hyperparameters. 

\subsubsection{Load-Balancing Penalty}
In multi-expert settings, routing collapse may occur if the deferral function over-relies on a subset of experts, leading to underuse or overload. To prevent this and encourage fair utilization, we introduce a \emph{Load-Balancing (LB) Penalty} that regulates the average pixel-wise routing allocation of each expert branch in a batch of samples.  
For each expert branch \(j \in \{1,\dots,J\}\), let the average routing ratio be
\begin{equation}
\rho_j = \frac{1}{BHW} \sum_{b=1}^{B} \sum_{i=1}^{HW} \mathcal{M}_j^{(b,i)},
\end{equation}
where \(B\) is the batch size, and \(\mathcal{M}_j^{(b,i)}\) is the routing probability of expert \(j\) for pixel \(i\) in the \(b\)-th image.

We impose lower and upper bounds $\mathbf{l} = [l_1,\dots,l_J]$ and $\mathbf{u} = [u_1,\dots,u_J]$ on the routing ratios, which are treated as configurable hyperparameters to avoid expert underuse or overload.
The penalty is defined as:
\begin{equation}
  \mathcal{L}_{\text{LB}} =
  \sum_{j=1}^{J}
      \left[ \max(\rho_j - u_j, 0) + \max(l_j - \rho_j, 0) \right].
  \label{eq:balance_multi}
\end{equation}

This penalty discourages extreme concentration or vanishing probability mass in the soft routing distribution and helps stabilize multi-expert training.
\subsubsection{The Learning Objective}
In the multi-expert setting, the training objective is:
\begin{equation}
  \mathcal{L}
  \;=\;
  \mathcal{L}_{\text{DC}}
  + \lambda_1\,\mathcal{L}_{\text{SC}}
  + \lambda_2\,\mathcal{L}_{\text{LB}} ,
  \label{eq:total_loss}
\end{equation}
where $\lambda_1$ and $\lambda_2$ are scalar weights that control the relative importance of spatial coherence and load balancing. For the setting of one expert, we can omit the $\mathcal{L}_{\text{LB}}$.


\begin{figure}[t]
\centering
\includegraphics[width=\linewidth]{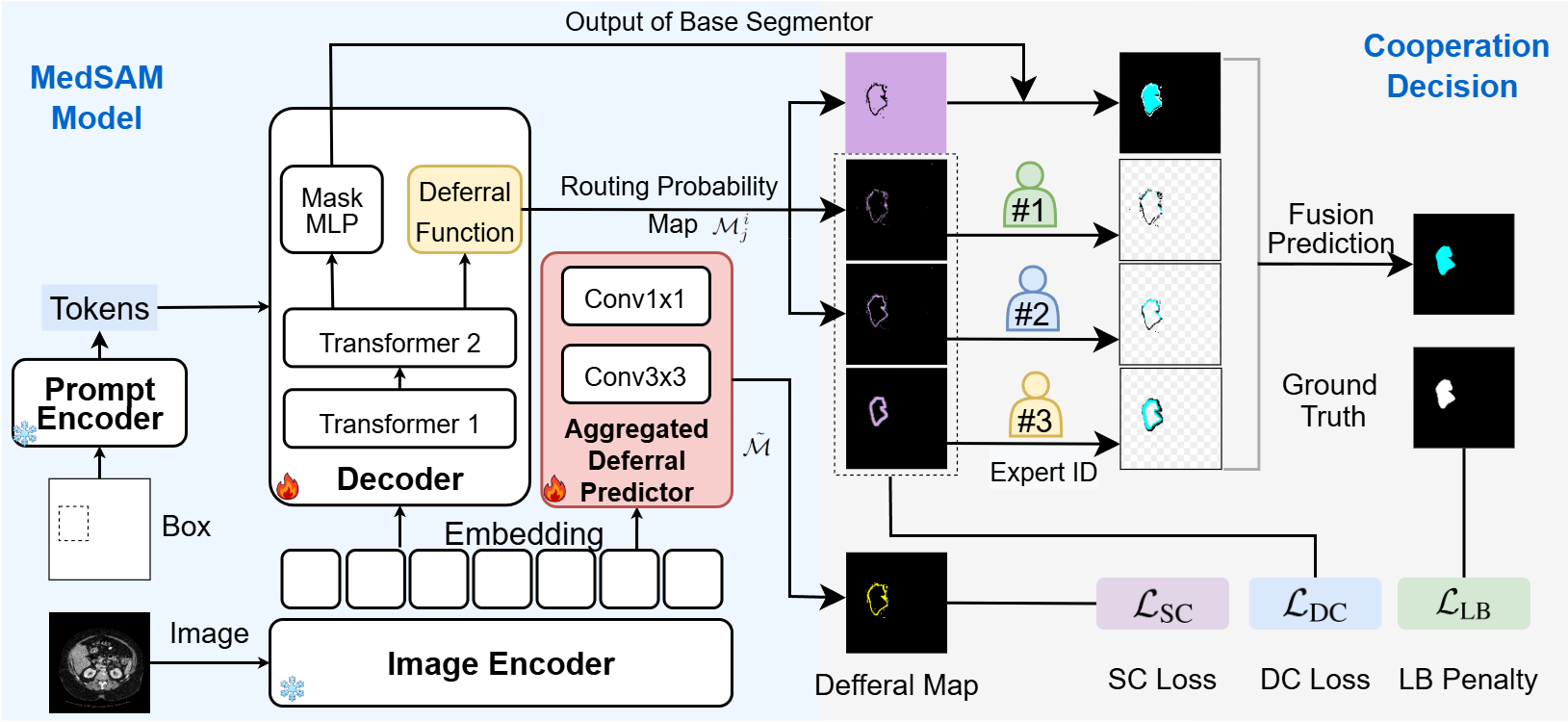}
\caption[Overall pipeline of the proposed framework.]{
Overall pipeline of the proposed $\nm$ framework, illustrated with a MedSAM-based instantiation.
The deferral function \(f_{\perp}\) denotes the routing head that outputs per-pixel routing logits.
The routing probability maps \(\mathcal{M}_j\) denote softmax-normalized branch-assignment probabilities.
The binary deferral mask \(\bar{\mathcal{M}}\) denotes the hard model/expert routing decision derived from the routing probability maps.
 The aggregated deferral predictor denotes the auxiliary CNN-based predictor that outputs \(\tilde{\mathcal{M}}\), a single-channel aggregated expert-assignment probability map used for spatial-coherence supervision.
}
\label{fig:framework}
\end{figure}

\subsubsection{Inference-time System Fusion}
\label{sec:system_fusion}

After training, the final System prediction is obtained by pixel-wise routing and fusion. For each pixel \(i\), the hard routing decision is determined by the largest routing logit:
\begin{equation}
r^i=\arg\max_{j\in\{0,\ldots,J\}} f_{\perp,j}(\vx^i),
\label{eq:routing_decision}
\end{equation}
where \(r^i=0\) indicates that pixel \(i\) is assigned to the model branch, and \(r^i>0\) indicates that it is assigned to one of the expert branches. The final System prediction is then obtained as
\begin{equation}
\hat{y}_{\mathrm{sys}}^{\,i} =
\begin{cases}
\tilde{y}^{\,i}, & r^i=0,\\
\mathbf{m}_{r^i}^{\,i}, & r^i\in\{1,\ldots,J\},
\end{cases}
\label{eq:system_fusion}
\end{equation}
where \(\tilde{y}^{\,i}=\mathbb{I}[\hat{y}^{\,i}\ge 0.5]\) is the binarized prediction of the base segmentor, and \(\mathbf{m}_{r^i}^{\,i}\) is the prediction from the selected expert branch. 
Therefore, the System prediction uses synthetic expert masks or clinician annotations only for pixels routed to expert branches, while pixels routed to the model branch are predicted by the base segmentor. 

\subsection{MedSAM-based Instantiation}\label{Incorporation}
$\nm$ is designed as a backbone-compatible deferral framework that can be integrated with segmentation models providing dense feature representations and pixel-wise predictions. In this work, we instantiate $\nm$ with MedSAM to leverage its strong cross-domain generalization for medical images, as illustrated in Fig.~\ref{fig:framework}.
Concretely, we keep the MedSAM encoder intact and extend the decoder by adding a deferral function and an aggregated deferral predictor, enabling pixel-wise routing to the base or expert branches without altering MedSAM's native mask outputs. This integration preserves compatibility with MedSAM while adding minimal overhead to support multi-expert collaboration.

\textbf{MedSAM Encoder:} The encoder of MedSAM remains frozen during training, ensuring that the model retains its cross-domain generalization ability. This frozen encoder processes input images to generate high-level feature representations that are then passed to the modified decoder.

\textbf{Modified Decoder:} 
To enable routing without altering MedSAM's native outputs, we augment its mask decoder with an additional deferral function. 
Concretely, MedSAM uses three learnable mask tokens and the same number of MLPs to produce native mask logits. 
We extend this design by adding $(J{+}1)$ \emph{routing tokens}: the first corresponds to the segmentor branch and the remaining $J$ correspond to the expert branches ($j{=}1,\dots,J$). 
In parallel, we expand the number of MLPs from size $3$ to $3+(J{+}1)$, where the first three MLPs are initialized from MedSAM's pretrained weights and fine-tuned during training, whereas the newly added $(J{+}1)$ heads are \emph{randomly initialized} to produce per-pixel routing logits.

\textbf{Aggregated Deferral Predictor:} 
To integrate it into MedSAM, we extend the mask decoder with an additional CNN-based aggregated deferral predictor. 
Concretely, given the decoder features $U\in\mathbb{R}^{C\times H\times W}$, 
the predictor applies a \(3\times 3\) convolution, ReLU activation, and a \(1\times 1\) convolution to produce a single-channel 
aggregated expert-assignment probability map, \(\tilde{\mathcal{M}} \in [0,1]^{H\times W}\).
This predictor runs in parallel with the existing routing logits $\{f_{\perp,j}\}_{j=0}^{J}$. 
While the routing logits are used for softmax-based deferral, the CNN predictor provides an explicit and spatially smooth estimate of the aggregated expert-assignment probability. 
During training, both outputs are supervised jointly by the spatial-coherence loss in Eq.~\eqref{eq:coherence_loss}. 
At inference time, the aggregated deferral predictor is used only for visualizing the overall deferral tendency and does not participate in the hard routing decision or final System fusion.

Through this MedSAM-based instantiation, $\nm$ leverages strong pretrained medical image representations while adding pixel-wise deferral capability for uncertain or challenging regions. Importantly, the same routing formulation can also be attached to other segmentation backbones, as further supported by the CENet transfer experiment.

\section{Experiments}\label{sec:experiments}

We conduct extensive experiments to evaluate $\nm$ from four perspectives: comparison with strong baselines, performance across organs and modalities, scalability to multiple experts, and ablation of key components.

\subsection{Datasets \& Metrics}
We evaluate the proposed method under two expert settings, as summarized in Tab.~\ref{tab:dataset-stats}: 
(1) \emph{synthetic-expert benchmarks}, including PROMISE12, LiTS, and AMOS22, where expert annotations are simulated to study controllable collaboration patterns; and 
(2) \emph{offline clinician-annotation benchmarks}, including Chaksu,
QUBIQ, and CURVAS, where pre-existing annotations from individual
clinicians are used to instantiate expert branches.

We report DSC, Jaccard, and Sensitivity in all experiments.

\subsection{Implementation Details}\label{sec:impl}

\subsubsection{Preprocessing}
For the MedSAM- and CENet-based experiments, 2D slices are resized to
\(1024\times1024\) and \(224\times224\), respectively. CT intensities are
clipped to \([-250,250]\) HU and normalized to \([0,1]\), while MRI volumes
are normalized using per-volume min--max scaling. Random flips and
\(90^\circ\) rotations are applied during training.

\subsubsection{Expert Sources}

The expert branch in $\nm$ conceptually represents the local correction
provided by a human expert after the model defers selected pixels or
regions. In the intended deployment, the learned deferral function first identifies these regions, and experts then provide corrections only within their assigned regions. In our experiments, synthetic expert masks and pre-existing clinician annotations serve as proxies for these prospective expert corrections. The former enables controlled analysis of
region-dependent expert reliability, while the latter preserves natural
inter-annotator variability. In both settings, the masks provide branch
predictions only for pixels selected by the learned routing function; they
are not used as routing inputs and do not determine where deferral occurs. These settings allow us to evaluate the routing behavior and System performance of $\nm$ when expert outputs are available.

\paragraph{Synthetic expert masks}
To reduce reliance on costly manual annotations or external experts, we propose the \texttt{SynthExpertSeg} function to instantiate \(J\) synthetic experts. Each expert \(j\in\{1,\dots,J\}\) is defined by an accuracy triplet ($[\mathrm{FG}_j,\ \mathrm{BG}_j,\ \mathrm{BD}_j]$), specifying its performance on foreground (FG), background (BG), and boundary (BD) regions. These synthetic experts are intentionally designed with imperfect and region-dependent reliability profiles, rather than as oracle experts. Therefore, the System cannot obtain a perfect mask simply by deferring more pixels; instead, it must learn whether the model branch or a given expert branch is more reliable for each pixel, which is the core setting where learning-to-defer is meaningful. The standard expert types and their accuracy settings used in our comparative experiments (SubSec.~\ref{Comparative_results}) are summarised in Tab.~\ref{tab:synthetic_experts}.

\begin{table*}[t]
  \centering

  \begin{minipage}[t]{0.70\textwidth}
    \centering
    \caption{Summary of the datasets.}
    \label{tab:dataset-stats}
    \vspace{1pt}
    \resizebox{\textwidth}{!}{%
    \begin{tabular}{lccc}
      \toprule
      \textbf{Dataset} &
      \textbf{Modality} &
      \textbf{\#Volumes/Images} &
      \textbf{ROI} \\
      \midrule
      PROMISE12~\citep{litjens2014evaluation} & MRI & 80 & Prostate \\
      LiTS~\citep{bilic2023liver} & CT & 201 & Liver / Tumor \\
      AMOS22~\citep{ji2022amos} & CT / MRI & 600 & 15 organs \\
      Chaksu~\citep{kumar2023chaksu} & Fundus & 1345 images & Optic Disc / Cup \\
      QUBIQ & MRI & 39 / 55 images & Brain / Prostate \\
      CURVAS & CT & 90 volumes & Kidney / Liver \\
      \bottomrule
    \end{tabular}%
    }
  \end{minipage}
  \hfill
  \begin{minipage}[t]{0.28\textwidth}
    \centering
    \caption{Region-specific accuracy profiles (\%) used to generate the synthetic expert masks in the comparative experiments.}
    \label{tab:synthetic_experts}
    \vspace{-6pt}
    \resizebox{\textwidth}{!}{%
    \begin{tabular}{c|ccc}
      \toprule
      \# \textbf{Expert} & \textbf{FG} & \textbf{BG} & \textbf{BD} \\
      \midrule
      1 & 92 & 98 & 98 \\
      2 & 85 & 99 & 94 \\
      3 & 75 & 95 & 90 \\
      \bottomrule
    \end{tabular}%
    }
  \end{minipage}
\end{table*}

\paragraph{Real clinician annotations}

For all offline clinician-annotation benchmarks, E1 is fixed as the
single expert branch in the one-expert setting, while E1, E2, and E3 are
used simultaneously in the three-expert setting. STAPLE (Simultaneous
Truth and Performance Level Estimation) is a probabilistic annotation-fusion
method that estimates a reference consensus mask from multiple clinician
annotations while accounting for their estimated annotation reliability.

For Chaksu, the released STAPLE consensus mask is used as the reference
mask, while the annotations from Expert 1, Expert 2, and Expert 3 are used
as E1, E2, and E3, respectively. For QUBIQ Brain Growth and QUBIQ Prostate,
the first three individual annotations are used as E1, E2, and E3, while
the remaining annotations are combined using STAPLE to construct the
reference masks. For CURVAS, the first three clinician annotations are
used as E1, E2, and E3, and their STAPLE consensus is used as the reference
mask. Therefore, the annotations used to construct the QUBIQ reference
masks are disjoint from those used as expert branches. In contrast, the
Chaksu and CURVAS reference masks are consensus estimates derived from the
same annotation pool used to instantiate the expert branches.

\subsubsection{Hyperparameter Configuration}
The overall training objective is defined in Eq.~\eqref{eq:total_loss}, with loss weights set to \(\lambda_1=1.0\) and \(\lambda_2=5.0\). Components are configured as: \textbf{SC Loss:} $\beta_1 = 0.5$ and $\beta_2 = 0.1$; \textbf{LB Penalty:} Upper and lower routing-allocation bounds for $J{=}3$ experts are \(\mathbf{u}=[0.35, 0.25, 0.30]\), \(\mathbf{l}=[0.05, 0.05, 0.05]\). Training details are provided in Appendix~\ref{app:implementation_details}.

\subsubsection{Hardware and Software Environment}
All experiments are conducted using Python 3.10, PyTorch 2.1, and MONAI 1.3 on NVIDIA RTX 4090 GPUs, with a single GPU used unless otherwise specified.

\subsection{Comparative Results}\label{Comparative_results}
\subsubsection{Quantitative Results}

We evaluate $\nm$ against strong automated baselines, including nnU-Net v2~\citep{isensee2024nnu}, CENet~\citep{bozorgpour2025cenet}, zero-shot MedSAM, and supervised fine-tuned MedSAM. 
These fully automatic baselines provide model-only reference performance, allowing us to quantify the potential benefit of pixel-wise deferral when most pixels are still handled by the model branch and only difficult pixels are deferred to expert branches. To make this additional expert cost explicit, we report the Model-only branch, the deferral ratio, and the expert routing allocation distribution in Tab.~\ref{tab:expert_workload}, showing that $\nm$ defers only a small fraction of pixels to expert branches.

We further include PatchCorr~\citep{koehler2024efficiently} as an expert-assisted patch-correction baseline. PatchCorr selects uncertain local patches according to the prediction-margin uncertainty of MedSAM and replaces the predictions within these patches using the same expert source as the one-expert setting of $\nm$. 
For a fair comparison of expert intervention, the correction budget of PatchCorr is matched to the one-expert deferral ratio of $\nm$ on each dataset.

\begin{table}[htbp]
  \centering
  \caption{Comparison across multiple segmentation tasks under
different evaluation and training settings (\%). The methods are grouped
into zero-shot inference, full task-specific training, expert-assisted
correction, and fine-tuned MedSAM-based comparison. MedSAM-FT (Fine-tuned) serves as
the direct backbone reference for DeferredSeg. Gray values following
$\uparrow$ denote absolute percentage-point improvements over MedSAM-FT.
Row-wise best values are shown in \textbf{bold}.}
  \label{tab:sota_summary}

  \setlength{\tabcolsep}{1.2pt}
  \renewcommand{\arraystretch}{0.82}
  \fontsize{8}{8.8}\selectfont

  \begin{adjustbox}{max width=\textwidth}
  \begin{tabular}{llcccccll}
  \toprule
  \multirow{2}{*}{\textbf{Dataset}}
  & \multirow{2}{*}{\textbf{Metric}}
  & \multicolumn{1}{c}{\shortstack{\textbf{Zero-shot}}}
  & \multicolumn{2}{c}{\shortstack{\textbf{Full task-specific}\\\textbf{training}}}
  & \multicolumn{1}{c}{\shortstack{\textbf{Expert-assisted}\\\textbf{correction}}}
  & \multicolumn{3}{c}{\shortstack{\textbf{Compare to the}\\\textbf{Fine-tuned}}} \\
  \cmidrule(lr){3-3}
  \cmidrule(lr){4-5}
  \cmidrule(lr){6-6}
  \cmidrule(lr){7-9}

  &
    & \multicolumn{1}{c}{\shortstack{\textbf{MedSAM}\\\textbf{ZS}}}
  & \multicolumn{1}{c}{\shortstack{\textbf{nnU-Net}\\\textbf{v2}}}
  & \multicolumn{1}{c}{\textbf{CENet}}
  & \multicolumn{1}{c}{\textbf{PatchCorr}}
  & \multicolumn{1}{c}{\shortstack{\textbf{MedSAM}\\\textbf{FT}}}
  & \multicolumn{1}{c}{\shortstack{\textbf{DeferredSeg}\\\textbf{1E}}}
  & \multicolumn{1}{c}{\shortstack{\textbf{DeferredSeg}\\\textbf{3E}}} \\
  \midrule

    \multirow{3}{*}{\shortstack{Prostate\\(PROMISE12)}}
      & DSC
        & 84.16 & 90.49 & 68.96 & 92.24 & 86.17
        & \textbf{97.06 }\gain{10.89}
        & 96.00 \gain{9.83} \\
      & Jaccard
        & 73.01 & 82.70 & 54.62 & 85.98 & 76.08
        & \textbf{94.38 }\gain{18.30}
        & 92.46 \gain{16.38} \\
      & Sens.
        & 76.70 & 91.78 & 75.20 & 94.45 & 84.97
        & 95.33 \gain{10.36}
        & \textbf{96.42 }\gain{11.45} \\
    \midrule

    \multirow{3}{*}{\shortstack{Liver\\(LiTS)}}
      & DSC
        & 91.64 & 96.62 & 93.61 & 92.86 & 91.69
        & \textbf{97.96 }\gain{6.27}
        & 96.15 \gain{4.46} \\
      & Jaccard
        & 86.43 & 93.48 & 89.52 & 88.83 & 86.49
        & \textbf{96.30 }\gain{9.81}
        & 93.70 \gain{7.21} \\
      & Sens.
        & 93.98 & \textbf{98.37} & 93.85 & 98.08 & 94.06
        & 96.83 \gain{2.77}
        & 95.47 \gain{1.41} \\
    \midrule

    \multirow{3}{*}{\shortstack{Tumor\\(LiTS)}}
      & DSC
        & 59.62 & 68.04 & 75.35 & 65.42 & 59.20
        & \textbf{91.01 }\gain{31.81}
        & 81.82 \gain{22.62} \\
      & Jaccard
        & 47.83 & 52.72 & 64.91 & 54.91 & 47.55
        & \textbf{85.81 }\gain{38.26}
        & 74.59 \gain{27.04} \\
      & Sens.
        & 76.18 & 84.99 & 76.38 & 87.75 & 75.97
        & \textbf{88.99 }\gain{13.02}
        & 84.39 \gain{8.42} \\
    \midrule

    \multirow{3}{*}{Spleen}
      & DSC
        & 89.21 & 94.29 & 89.62 & 93.80 & 94.57
        & 97.75 \gain{3.18}
        & \textbf{98.14 }\gain{3.57} \\
      & Jaccard
        & 81.74 & 90.83 & 84.79 & 89.33 & 90.77
        & 95.86 \gain{5.09}
        & \textbf{96.59 }\gain{5.82} \\
      & Sens.
        & 90.78 & 95.21 & 89.41 & 95.41 & 95.02
        & 98.06 \gain{3.04}
        & \textbf{98.34 }\gain{3.32} \\
    \midrule

    \multirow{3}{*}{Right Kidney}
      & DSC
        & 89.39 & 90.26 & 91.34 & 93.76 & 94.83
        & \textbf{98.34 }\gain{3.51}
        & \textbf{98.34 }\gain{3.51} \\
      & Jaccard
        & 81.98 & 85.99 & 86.47 & 89.05 & 90.85
        & \textbf{96.85 }\gain{6.00}
        & 96.84 \gain{5.99} \\
      & Sens.
        & 91.70 & 85.86 & 91.27 & 96.46 & 95.25
        & \textbf{98.69 }\gain{3.44}
        & 98.49 \gain{3.24} \\
    \midrule

    \multirow{3}{*}{Left Kidney}
      & DSC
        & 89.63 & 93.67 & 91.61 & 93.42 & 94.74
        & 98.14 \gain{3.40}
        & \textbf{98.16 }\gain{3.42} \\
      & Jaccard
        & 82.33 & 88.99 & 86.92 & 88.63 & 90.86
        & 96.60 \gain{5.74}
        & \textbf{96.63 }\gain{5.77} \\
      & Sens.
        & 93.11 & 93.21 & 91.39 & 97.62 & 94.67
        & \textbf{98.72 }\gain{4.05}
        & 98.25 \gain{3.58} \\
    \midrule

    \multirow{3}{*}{Gallbladder}
      & DSC
        & 79.60 & 76.73 & 62.58 & 90.82 & 86.69
        & \textbf{96.57 }\gain{9.88}
        & 95.27 \gain{8.58} \\
      & Jaccard
        & 68.58 & 68.31 & 45.66 & 84.62 & 78.77
        & \textbf{93.88 }\gain{15.11}
        & 91.82 \gain{13.05} \\
      & Sens.
        & 79.56 & 81.95 & 50.48 & 92.20 & 88.46
        & \textbf{97.27 }\gain{8.81}
        & 96.62 \gain{8.16} \\
    \midrule

    \multirow{3}{*}{Esophagus}
      & DSC
        & 65.72 & 82.71 & 66.44 & 90.96 & 79.67
        & 96.14 \gain{16.47}
        & \textbf{96.44 }\gain{16.77} \\
      & Jaccard
        & 51.50 & 72.07 & 55.24 & 84.10 & 68.37
        & 92.78 \gain{24.41}
        & \textbf{93.28 }\gain{24.91}\\
      & Sens.
        & 67.36 & 84.43 & 65.63 & 88.46 & 78.58
        & \textbf{96.50 }\gain{17.92}
        & 96.43 \gain{17.85} \\
    \midrule

    \multirow{3}{*}{\shortstack{Liver\\(AMOS22)}}
      & DSC
        & 86.78 & 97.17 & 90.90 & 89.29 & 95.15
        & 97.54 \gain{2.39}
        & \textbf{97.87 }\gain{2.72} \\
      & Jaccard
        & 78.67 & 94.57 & 86.80 & 82.85 & 91.82
        & 95.48 \gain{3.66}
        & \textbf{96.07 }\gain{4.25} \\
      & Sens.
        & 91.25 & 97.37 & 90.64 & 95.99 & 95.44
        & 98.44 \gain{3.00}
        & \textbf{98.70 }\gain{3.26} \\
    \midrule

    \multirow{3}{*}{Stomach}
      & DSC
        & 81.56 & 87.27 & 77.07 & 89.81 & 90.69
        & \textbf{96.70 }\gain{6.01}
        & 96.67 \gain{5.98} \\
      & Jaccard
        & 70.82 & 80.14 & 70.37 & 83.32 & 84.98
        & \textbf{94.03 }\gain{9.05}
        & 93.98 \gain{9.00} \\
      & Sens.
        & 81.04 & 89.78 & 76.35 & 92.84 & 90.46
        & \textbf{97.62 }\gain{7.16}
        & \textbf{97.62 }\gain{7.16} \\
    \midrule

    \multirow{3}{*}{Aorta}
      & DSC
        & 86.45 & 94.85 & 86.35 & 93.57 & 94.61
        & 98.00 \gain{3.39}
        & \textbf{98.20 }\gain{3.59} \\
      & Jaccard
        & 78.02 & 90.40 & 80.95 & 89.90 & 90.36
        & 96.34 \gain{5.98}
        & \textbf{96.68 }\gain{6.32} \\
      & Sens.
        & 87.59 & 93.75 & 85.29 & 94.69 & 94.35
        & 98.11 \gain{3.76}
        & \textbf{98.47 }\gain{4.12} \\
    \midrule

    \multirow{3}{*}{\shortstack{Inferior\\vena cava}}
      & DSC
        & 77.59 & 87.84 & 75.42 & 92.37 & 85.42
        & \textbf{97.02 }\gain{11.60}
        & 97.01 \gain{11.59} \\
      & Jaccard
        & 65.58 & 78.84 & 66.16 & 86.43 & 76.61
        & \textbf{94.39 }\gain{17.78}
        & 94.38 \gain{17.77} \\
      & Sens.
        & 75.48 & 88.42 & 85.20 & 89.20 & 83.88
        & 97.17 \gain{13.29}
        & \textbf{97.23 }\gain{13.35} \\
    \midrule

    \multirow{3}{*}{Pancreas}
      & DSC
        & 69.30 & 82.06 & 62.46 & 80.51 & 83.90
        & \textbf{95.68 }\gain{11.78}
        & 95.39 \gain{11.49} \\
      & Jaccard
        & 56.06 & 71.11 & 47.18 & 71.06 & 74.11
        & \textbf{92.19 }\gain{18.08}
        & 91.67 \gain{17.56} \\
      & Sens.
        & 77.23 & 85.44 & 56.61 & 90.46 & 86.92
        & \textbf{96.54 }\gain{9.62}
        & 95.82 \gain{8.90} \\
    \midrule

    \multirow{3}{*}{Duodenum}
      & DSC
        & 62.14 & 74.75 & 38.54 & 75.44 & 79.83
        & 94.63 \gain{14.80}
        & \textbf{94.86 }\gain{15.03} \\
      & Jaccard
        & 48.34 & 62.07 & 25.74 & 65.21 & 68.50
        & 90.38 \gain{21.88}
        & \textbf{90.72 }\gain{22.22} \\
      & Sens.
        & 69.37 & 74.59 & 30.89 & 82.02 & 81.53
        & \textbf{95.40 }\gain{13.87}
        & 95.38 \gain{13.85} \\
    \midrule

    \multirow{3}{*}{Bladder}
      & DSC
        & 81.64 & 87.11 & 75.38 & 89.50 & 89.17
        & \textbf{95.99 }\gain{6.82}
        & \textbf{95.99 }\gain{6.82} \\
      & Jaccard
        & 71.80 & 80.74 & 61.78 & 83.42 & 82.72
        & 93.27 \gain{10.55}
        & \textbf{93.29 }\gain{10.57} \\
      & Sens.
        & 81.94 & 94.39 & 69.68 & 93.21 & 92.45
        & \textbf{97.68 }\gain{5.23}
        & 97.62 \gain{5.17} \\
    \midrule

    \multirow{3}{*}{\shortstack{Prostate/\\Uterus}}
      & DSC
        & 81.11 & 85.40 & 41.64 & 93.16 & 88.73
        & 97.09 \gain{8.36}
        & \textbf{97.19 }\gain{8.46} \\
      & Jaccard
        & 69.65 & 75.83 & 29.57 & 87.77 & 80.77
        & 94.58 \gain{13.81}
        & \textbf{94.78 }\gain{14.01}\\
      & Sens.
        & 76.65 & 88.46 & 36.72 & 92.42 & 91.44
        & 97.21 \gain{5.77}
        & \textbf{97.26 }\gain{5.82} \\
    \bottomrule
  \end{tabular}
  \end{adjustbox}
\end{table}

\noindent\textbf{Synthetic-expert evaluation.}

The main quantitative comparison is summarized in Tab.~\ref{tab:sota_summary}, while the complete results with standard deviations and the Expert/Model branches are provided in~\ref{sec:appendix_sota_summary}. For $\nm$, we report three perspectives: \emph{System} performance, \emph{Expert-only} performance, and \emph{Model-only} performance, following the inference-time deferral rule defined in Eq.~\eqref{eq:system_fusion}.
The computational overhead of introducing the deferral mechanism is
also evaluated in ~\ref{sec:computational_overhead}, showing
only a small increase in model complexity and inference cost over the
base MedSAM.

Across PROMISE12, LiTS, and AMOS22, $\nm$ achieves improved
System-level performance over the automated baselines in most settings.
PROMISE12 evaluates a single mid-sized organ with ambiguous gland
boundaries, where both the one- and three-expert configurations improve
overlap and sensitivity by selectively routing difficult boundary pixels. Additional leave-one-center-out validation on PROMISE12 further confirms
that $\nm$, particularly with three experts, improves overlap-based
performance under center shifts with limited expert deferral
(Appendix~\ref{app:cross_center}).
LiTS contains large homogeneous liver regions together with small,
low-contrast tumors; $\nm$ improves both tasks, with particularly clear
gains for tumor segmentation, while the three-expert setting remains
competitive for liver segmentation. AMOS22 further evaluates 13 abdominal
organs with diverse sizes and contrasts. On this multi-organ benchmark,
$\nm$ performs strongly against nnU-Net v2, CENet, and MedSAM, with
statistically significant improvements over MedSAM for most organs
(Tab.~\ref{tab:significance_summary}), especially for small,
low-contrast, and anatomically complex structures. 

It should be noted that the expert branches are not assumed to be globally superior to the model branch. As shown in the complete branch-level results, in several cases such as Left kidney, Spleen, and Inferior vena cava, the Expert-only branch is even weaker than the Model-only branch. This indicates that the gain of $\nm$ does not come from simply replacing the model prediction with stronger expert predictions. Instead, $\nm$ learns to selectively defer pixels or regions to expert branches only when they are locally more reliable, while retaining the model prediction in regions where the model branch is more competent.

\noindent\textbf{Real-expert evaluation.}

In this subsection, the expert branches are instantiated using clinician
annotations. The main quantitative comparison is summarized in
Tab.~\ref{tab:real_expert_summary}, while the complete results with
standard deviations and the Expert/Model branches are provided in
~\ref{sec:appendix_real_expert}.

\begin{table}[htbp]
  \centering
  \caption{Real-expert evaluation across Chaksu, QUBIQ, and CURVAS (\%). MedSAM-FT serves as the direct backbone reference for DeferredSeg. Gray values following the arrows denote absolute percentage-point changes relative to MedSAM-FT. Row-wise best values are shown in \textbf{bold}.}
  \label{tab:real_expert_summary}

  \setlength{\tabcolsep}{1.8pt}
  \renewcommand{\arraystretch}{0.78}
  \fontsize{10}{10.5}\selectfont

  \begin{adjustbox}{max width=\textwidth}
  \begin{tabular}{llccccll}
    \toprule
    \multirow{2}{*}{\textbf{Dataset}}
    & \multirow{2}{*}{\textbf{Metric}}
    & \multicolumn{1}{c}{\shortstack{\textbf{Zero-shot}}}
    & \multicolumn{1}{c}{\shortstack{\textbf{Task-specific}\\\textbf{training}}}
    & \multicolumn{1}{c}{\shortstack{\textbf{Expert-assisted}\\\textbf{correction}}}
    & \multicolumn{3}{c}{\shortstack{\textbf{Compare to the}\\\textbf{Fine-tuned}}} \\
    \cmidrule(lr){3-3}
    \cmidrule(lr){4-4}
    \cmidrule(lr){5-5}
    \cmidrule(lr){6-8}

    &
       & \multicolumn{1}{c}{\shortstack{\textbf{MedSAM}\\\textbf{ZS}}}
    & \multicolumn{1}{c}{\textbf{CENet}}
    & \multicolumn{1}{c}{\textbf{PatchCorr}}
    & \multicolumn{1}{c}{\shortstack{\textbf{MedSAM}\\\textbf{FT}}}
    & \multicolumn{1}{c}{\shortstack{\textbf{DeferredSeg}\\\textbf{1E}}}
    & \multicolumn{1}{c}{\shortstack{\textbf{DeferredSeg}\\\textbf{3E}}} \\
    \midrule

    \multirow{3}{*}{Chaksu}
    & DSC
    & 94.97
    & 97.20
    & 90.37
    & 96.76
    & \textbf{97.36}\gain{0.60}
    & 97.17\gain{0.41} \\

    & Jaccard
    & 90.68
    & 94.64
    & 82.78
    & 93.89
    & \textbf{94.87}\gain{0.98}
    & 94.60\gain{0.71} \\

    & Sens.
    & 91.51
    & 96.67
    & 84.11
    & 96.69
    & 96.30\drop{0.39}
    & \textbf{96.74}\gain{0.05} \\
    \midrule

    \multirow{3}{*}{\shortstack{Brain Growth\\(QUBIQ)}}
    & DSC
    & 66.98
    & 86.76
    & 82.10
    & 83.15
    & 86.68\gain{3.53}
    & \textbf{89.33}\gain{6.18} \\

    & Jaccard
    & 50.57
    & 79.10
    & 72.37
    & 71.28
    & 78.52\gain{7.24}
    & \textbf{81.51}\gain{10.23} \\

    & Sens.
    & 80.86
    & 86.04
    & 84.38
    & 85.03
    & 86.49\gain{1.46}
    & \textbf{88.20}\gain{3.17} \\
    \midrule

    \multirow{3}{*}{\shortstack{Prostate\\(QUBIQ)}}
    & DSC
    & 93.39
    & 94.23
    & 95.62
    & 94.77
    & \textbf{96.73}\gain{1.96}
    & 95.71\gain{0.94} \\

    & Jaccard
    & 89.24
    & 89.36
    & 91.68
    & 90.16
    & \textbf{94.56}\gain{4.40}
    & 92.17\gain{2.01} \\

    & Sens.
    & 92.16
    & \textbf{96.72}
    & 96.29
    & 96.20
    & 93.69\drop{2.51}
    & 92.76\drop{3.44} \\
    \midrule

    \multirow{3}{*}{\shortstack{Kidney\\(CURVAS)}}
    & DSC
    & 67.94
    & 93.17
    & 91.32
    & 90.70
    & 94.48\gain{3.78}
    & \textbf{95.80}\gain{5.10} \\

    & Jaccard
    & 54.93
    & 88.78
    & 86.75
    & 84.41
    & 90.00\gain{5.59}
    & \textbf{91.98}\gain{7.57} \\

    & Sens.
    & 88.90
    & 92.45
    & 89.49
    & 91.83
    & 92.67\gain{0.84}
    & \textbf{93.97}\gain{2.14} \\
    \midrule

    \multirow{3}{*}{\shortstack{Liver\\(CURVAS)}}
    & DSC
    & 92.51
    & 93.84
    & 92.39
    & 93.14
    & 93.55\gain{0.41}
    & \textbf{95.80}\gain{2.66} \\

    & Jaccard
    & 87.74
    & 88.43
    & 87.24
    & 88.39
    & 89.93\gain{1.54}
    & \textbf{91.94}\gain{3.55} \\

    & Sens.
    & 92.68
    & 93.65
    & 92.16
    & 94.17
    & \textbf{94.88}\gain{0.71}
    & 94.04\drop{0.13} \\

    \bottomrule
  \end{tabular}
  \end{adjustbox}
\end{table}

Across the three clinician-annotation benchmarks, the benefit of
multi-expert deferral depends on task-specific expert complementarity.
For Chaksu, where both the model and clinician-annotation branches are
already strong, the one-expert configuration achieves the best
overlap-based performance, while the multi-expert setting remains
competitive and outperforms PatchCorr. On QUBIQ, the benefit of multiple experts is task-dependent: they improve
brain-growth segmentation, whereas the one-expert setting performs best
for prostate segmentation. Notably, for prostate segmentation, $\nm$
still benefits from selectively using locally reliable expert regions
even when the Expert-only branch is weaker than the model branch. On CURVAS, multi-expert collaboration is particularly
beneficial for kidney segmentation, whereas liver segmentation presents a
metric-dependent trade-off between the one- and three-expert settings.
These results show that $\nm$ can exploit locally reliable expert regions,
but that the value of adding experts depends on their complementarity and
the target task.

\noindent\textbf{Expert routing allocation and deferral ratio.}

Tab.~\ref{tab:expert_workload} reports the routing distribution. Across both synthetic- and real-expert benchmarks, the model branch processes the large majority of pixels, while only a limited subset is deferred to expert branches. The multi-expert setting generally increases the total deferral ratio, particularly on PROMISE12, QUBIQ-Prostate, and the CURVAS tasks, but most pixels remain assigned to the model branch. The individual expert routing allocations are not necessarily uniform because the routing function selects expert branches according to their local pixel-level reliability. These results show that the System improvements are achieved through selective pixel-wise deferral rather than full-image expert intervention.

\begin{table}[!htbp]
\centering
\caption{Expert routing allocation and deferral ratios under the one- and
three-expert settings for synthetic expert masks and clinician
annotations. Values are reported as percentages. Deferred denotes the
percentage of pixels routed to expert branches, and Model denotes the
percentage retained by the model branch.}
\label{tab:expert_workload}

\setlength{\tabcolsep}{1.8pt}
\renewcommand{\arraystretch}{0.84}
\fontsize{8}{9}\selectfont

\begin{adjustbox}{max width=\textwidth}
\begin{tabular}{@{}llccc|ccc@{}}
\toprule
\multirow{2}{*}{\textbf{Expert source}}
& \multirow{2}{*}{\textbf{Dataset}}
& \multicolumn{3}{c|}{\textbf{One expert}}
& \multicolumn{3}{c}{\textbf{Three experts}} \\
\cmidrule(lr){3-5}
\cmidrule(lr){6-8}

&
& \textbf{Deferred}
& \textbf{E1}
& \textbf{Model}
& \textbf{Deferred}
& \textbf{E1 / E2 / E3}
& \textbf{Model} \\
\midrule

\multirow{16}{*}{\shortstack{Synthetic\\expert masks}}
& PROMISE12
& 4.32 & 4.32 & 95.68
& 10.28 & 0.80 / 8.32 / 1.16 & 89.72 \\

& LiTS-Liver
& 1.57 & 1.57 & 98.43
& 1.45 & 1.45 / $<0.01$ / $<0.01$ & 98.55 \\

& LiTS-Tumor
& 0.48 & 0.48 & 99.52
& 0.46 & 0.40 / 0.04 / 0.02 & 99.54 \\

& Spleen
& 0.55 & 0.55 & 99.45
& 0.72 & 0.70 / 0.01 / 0.01 & 99.28 \\

& Right Kidney
& 0.48 & 0.48 & 99.52
& 0.66 & 0.62 / 0.03 / 0.01 & 99.34 \\

& Left Kidney
& 0.53 & 0.53 & 99.47
& 0.67 & 0.62 / 0.04 / 0.01 & 99.33 \\

& Gallbladder
& 0.38 & 0.38 & 99.62
& 2.57 & 0.59 / 1.98 / 0.00 & 97.43 \\

& Esophagus
& 0.11 & 0.11 & 99.89
& 0.19 & 0.11 / 0.07 / 0.01 & 99.81 \\

& Liver
& 1.75 & 1.75 & 98.25
& 3.02 & 2.01 / 1.00 / $<0.01$ & 96.98 \\

& Stomach
& 1.29 & 1.29 & 98.71
& 2.23 & 1.32 / 0.88 / 0.02 & 97.77 \\

& Aorta
& 0.19 & 0.19 & 99.81
& 1.67 & 0.59 / 1.00 / 0.08 & 98.33 \\

& \shortstack[l]{Inferior vena cava}
& 0.17 & 0.17 & 99.83
& 0.25 & 0.18 / 0.07 / 0.00 & 99.75 \\

& Pancreas
& 0.72 & 0.72 & 99.28
& 0.43 & 0.28 / 0.11 / 0.04 & 99.57 \\

& Duodenum
& 0.45 & 0.45 & 99.55
& 0.50 & 0.40 / 0.07 / 0.02 & 99.50 \\

& Bladder
& 1.04 & 1.04 & 98.96
& 1.03 & 0.89 / 0.13 / 0.01 & 98.97 \\

& \shortstack[l]{Prostate/Uterus}
& 0.61 & 0.61 & 99.39
& 0.49 & 0.49 / 0.00 / 0.00 & 99.51 \\

\midrule

\multirow{5}{*}{\shortstack{Clinician\\annotations}}
& Chaksu
& 1.98 & 1.98 & 98.02
& 3.11 & 1.02 / 1.11 / 0.98 & 96.89 \\

& QUBIQ-Brain Growth
& 1.27 & 1.27 & 98.73
& 3.37 & 1.41 / 0.97 / 0.98 & 96.63 \\

& QUBIQ-Prostate
& 1.28 & 1.28 & 98.72
& 7.88 & 2.97 / 1.91 / 3.00 & 92.12 \\

& CURVAS-Kidney
& 3.22 & 3.22 & 96.78
& 5.94 & 1.41 / 2.53 / 2.00 & 94.06 \\

& CURVAS-Liver
& 1.63 & 1.63 & 98.37
& 5.32 & 1.69 / 2.27 / 1.36 & 94.68 \\

\bottomrule
\end{tabular}
\end{adjustbox}
\end{table}

\noindent\textbf{Transferability to fully supervised CENet (Tab.~\ref{tab:cenet_l2d_3exp}).}

To verify backbone compatibility, we integrate $\nm$ into CENet under the same three-expert setting. Although CENet already provides a strong baseline, adding $\nm$ brings further gains across several synthetic-expert benchmarks and remains competitive on the real-expert Chaksu benchmark. These results indicate that the proposed deferral mechanism is not specific to MedSAM and can generalize to different segmentation backbones.

\begin{table}[t]
  \centering
  \caption{CENet baseline versus CENet with L2D under the three-expert setting
  (mean $\pm$ standard deviation, \%). Row-wise best values are
  \textbf{bold}.}
  
  \label{tab:cenet_l2d_3exp}

  \renewcommand{\arraystretch}{1.0}
  \setlength{\tabcolsep}{4pt}
  \fontsize{8}{8.5}\selectfont

  \begin{tabular}{llc|ccc}
    \toprule
    \textbf{Dataset}
    & \textbf{Metric}
    & \textbf{CENet}
    & \multicolumn{3}{c}{\textbf{Ours (3 experts)}} \\
    \cmidrule(lr){4-6}

    &
    & \textbf{}
    & \textbf{System}
    & \textbf{Expert}
    & \textbf{Model} \\
    \midrule
\multirow{3}{*}{\shortstack{Prostate\\(PROMISE12)}}
  & DSC     
    & 68.96 $\pm$ 0.16  & \textbf{88.85 $\pm$ 0.07}  & \textcolor{gray}{89.80 $\pm$ 0.82} & \textcolor{gray}{71.98 $\pm$ 6.69} \\
  & Jaccard 
    & 54.62 $\pm$ 0.13 & \textbf{80.47 $\pm$ 0.03}  & \textcolor{gray}{82.01 $\pm$ 1.22} & \textcolor{gray}{59.89 $\pm$ 3.87} \\
  & Sens.   
    & 75.20 $\pm$ 0.10   & \textbf{92.79 $\pm$ 2.38}  & \textcolor{gray}{94.24 $\pm$ 0.87} & \textcolor{gray}{72.41 $\pm$ 7.32} \\
\midrule

\multirow{3}{*}{\shortstack{Liver\\(LiTS)}}
  & DSC     & 93.61 $\pm$ 0.10 & \textbf{95.16 $\pm$ 0.48} & \textcolor{gray}{94.19 $\pm$ 1.00} & \textcolor{gray}{94.75 $\pm$ 1.86} \\
  & Jaccard & 89.52 $\pm$ 0.07 
            & \textbf{91.73 $\pm$ 0.71} & \textcolor{gray}{89.04 $\pm$ 0.97} & \textcolor{gray}{91.95 $\pm$ 2.70} \\
  & Sens.   & 93.85 $\pm$ 0.14 & \textbf{94.03 $\pm$ 0.71} & \textcolor{gray}{95.67 $\pm$ 0.71} & \textcolor{gray}{95.57 $\pm$ 1.44} \\
\midrule
\multirow{3}{*}{\shortstack{Tumor\\(LiTS)}}
  & DSC     & 75.35 $\pm$ 0.52 & \textbf{79.06 $\pm$ 0.14} & \textcolor{gray}{84.42 $\pm$ 0.22} & \textcolor{gray}{69.49 $\pm$ 0.31} \\
  & Jaccard & 64.91 $\pm$ 0.36 & \textbf{65.37 $\pm$ 0.17} & \textcolor{gray}{76.08 $\pm$ 0.25} & \textcolor{gray}{53.24 $\pm$ 0.29} \\
  & Sens.   & 76.38 $\pm$ 0.65 & \textbf{78.57 $\pm$ 0.11} & \textcolor{gray}{79.03 $\pm$ 0.19} & \textcolor{gray}{59.62 $\pm$ 0.27} \\
\midrule

\multirow{3}{*}{Spleen}
  & DSC     & 89.62 $\pm$ 0.25 & \textbf{93.94 $\pm$ 0.09} & \textcolor{gray}{90.41 $\pm$ 0.05} & \textcolor{gray}{89.52 $\pm$ 2.46} \\
  & Jaccard & 84.79 $\pm$ 0.26 & \textbf{88.58 $\pm$ 0.16} & \textcolor{gray}{82.55 $\pm$ 0.08} & \textcolor{gray}{85.56 $\pm$ 1.20} \\
  & Sens.   & 89.41 $\pm$ 0.28 & \textbf{90.89 $\pm$ 0.01} & \textcolor{gray}{85.96 $\pm$ 0.10} & \textcolor{gray}{88.60 $\pm$ 3.11} \\
\midrule
\multirow{3}{*}{Right Kidney}
  & DSC     & 91.34 $\pm$ 0.86 & \textbf{95.35 $\pm$ 0.12} & \textcolor{gray}{88.31 $\pm$ 0.05} & \textcolor{gray}{91.67 $\pm$ 0.15} \\
  & Jaccard & 86.47 $\pm$ 0.93 & \textbf{91.11 $\pm$ 0.08} & \textcolor{gray}{79.26 $\pm$ 0.08} & \textcolor{gray}{87.65 $\pm$ 0.08} \\
  & Sens.   & 91.27 $\pm$ 1.16 & \textbf{94.00 $\pm$ 0.06} & \textcolor{gray}{81.20 $\pm$ 0.01} & \textcolor{gray}{91.31 $\pm$ 0.97} \\
\midrule
\multirow{3}{*}{\shortstack{Chaksu}}
  & DSC     
    & 97.20 $\pm$ 0.06 & \textbf{97.27 $\pm$ 0.17} & \textcolor{gray}{97.32 $\pm$ 0.15} & \textcolor{gray}{97.19 $\pm$ 0.21} \\
  & Jaccard 
    & 94.64 $\pm$ 0.08 & \textbf{94.74 $\pm$ 0.30} & \textcolor{gray}{94.84 $\pm$ 0.27} & \textcolor{gray}{94.60 $\pm$ 0.39} \\
  & Sens.   
    & \textbf{96.67 $\pm$ 0.15} & 95.87 $\pm$ 0.52 & \textcolor{gray}{95.94 $\pm$ 0.51} & \textcolor{gray}{95.82 $\pm$ 0.52} \\
    \bottomrule
  \end{tabular}
\end{table}

\noindent\textbf{Statistical significance analysis.}

We conduct one-sided paired Wilcoxon signed-rank tests on case-level
DSC scores. The MedSAM-based one- and three-expert configurations are
compared with MedSAM-FT and PatchCorr, while the CENet-based configuration
is compared with CENet-FT under the same backbone. When multiple random
seeds are available, per-case DSC scores are first averaged across seeds.
For a baseline \(b\), the paired difference is defined as
$d_i^{(b)}=\mathrm{DSC}_{\nm,i}-\mathrm{DSC}_{b,i}$.
The null hypothesis assumes no positive shift in the paired differences,
whereas the alternative hypothesis assumes a positive shift, indicating
that $\nm$ achieves higher case-level DSC than the corresponding baseline.

\begin{table*}[t]
\centering
\begin{minipage}[t]{0.46\textwidth}
\centering

\caption{Summary of statistically significant improvements in
case-level DSC. Values denote the number of significant comparisons over
the total number of comparisons. Detailed results are provided in
~\ref{app:statistical_significance}.}
\label{tab:significance_summary}

\setlength{\tabcolsep}{5pt}
\renewcommand{\arraystretch}{0.95}
\fontsize{8pt}{8.6pt}\selectfont

\begin{tabular}{lcc}
\toprule
Baseline
& 1E
& 3E\\
\midrule
MedSAM-FT
& 15/16
& 15/16\\

PatchCorr
& 16/16
& 16/16\\

CENet-FT
& --
& 6/6\\
\bottomrule
\end{tabular}

\end{minipage}
\hfill
\begin{minipage}[t]{0.50\textwidth}
\centering

\caption{Analysis of each loss component through ablation. We report the
overall System performance in the main text; detailed Expert and Model
results are provided in the appendix Table~\ref{tab:ablation_all}.}
\label{tab:ablation}

\setlength{\tabcolsep}{4pt}
\renewcommand{\arraystretch}{0.9}
\fontsize{8pt}{8.6pt}\selectfont

\begin{tabular}{llccc}
\toprule
\textbf{J} & \textbf{Loss}
& \textbf{DSC}
& \textbf{Jaccard}
& \textbf{Sens.} \\
\midrule
\multirow{2}{*}{1}
  & DC
    & 96.86$\pm$0.92
    & 94.00$\pm$1.73
    & 96.80$\pm$1.89 \\
  & + SC
    & \textbf{97.17$\pm$0.63}
    & \textbf{94.57$\pm$1.17}
    & \textbf{97.29$\pm$1.24} \\
\midrule
\multirow{4}{*}{3}
  & DC
    & 95.39$\pm$1.64
    & 91.51$\pm$2.95
    & 97.13$\pm$2.23 \\
  & + SC
    & 95.79$\pm$1.71
    & 92.14$\pm$3.07
    & 97.18$\pm$2.31 \\
  & + LB
    & \textbf{96.59$\pm$1.46}
    & 93.52$\pm$2.69
    & 97.16$\pm$2.19 \\
  & + SC, LB
    & 96.58$\pm$1.04
    & \textbf{93.56$\pm$1.87}
    & \textbf{98.34$\pm$0.47} \\
\bottomrule
\end{tabular}

\end{minipage}

\end{table*}

As summarized in Tab.~\ref{tab:significance_summary}, $\nm$ achieves
significant improvements over MedSAM-FT in 30 of 32 comparisons and over
PatchCorr in all 32 comparisons. Under the CENet backbone, $\nm$ achieves statistically significant improvements over CENet-FT on all six evaluated segmentation tasks.

\subsubsection{Qualitative Results}
We present per-slice comparisons and deferral heatmaps to visualize
System predictions and expert-assignment probabilities.

\begin{figure}[!t]
  \centering
  \begin{minipage}[t]{0.76\linewidth}
  \vspace{0pt}
    \centering
    \includegraphics[width=0.98\linewidth]{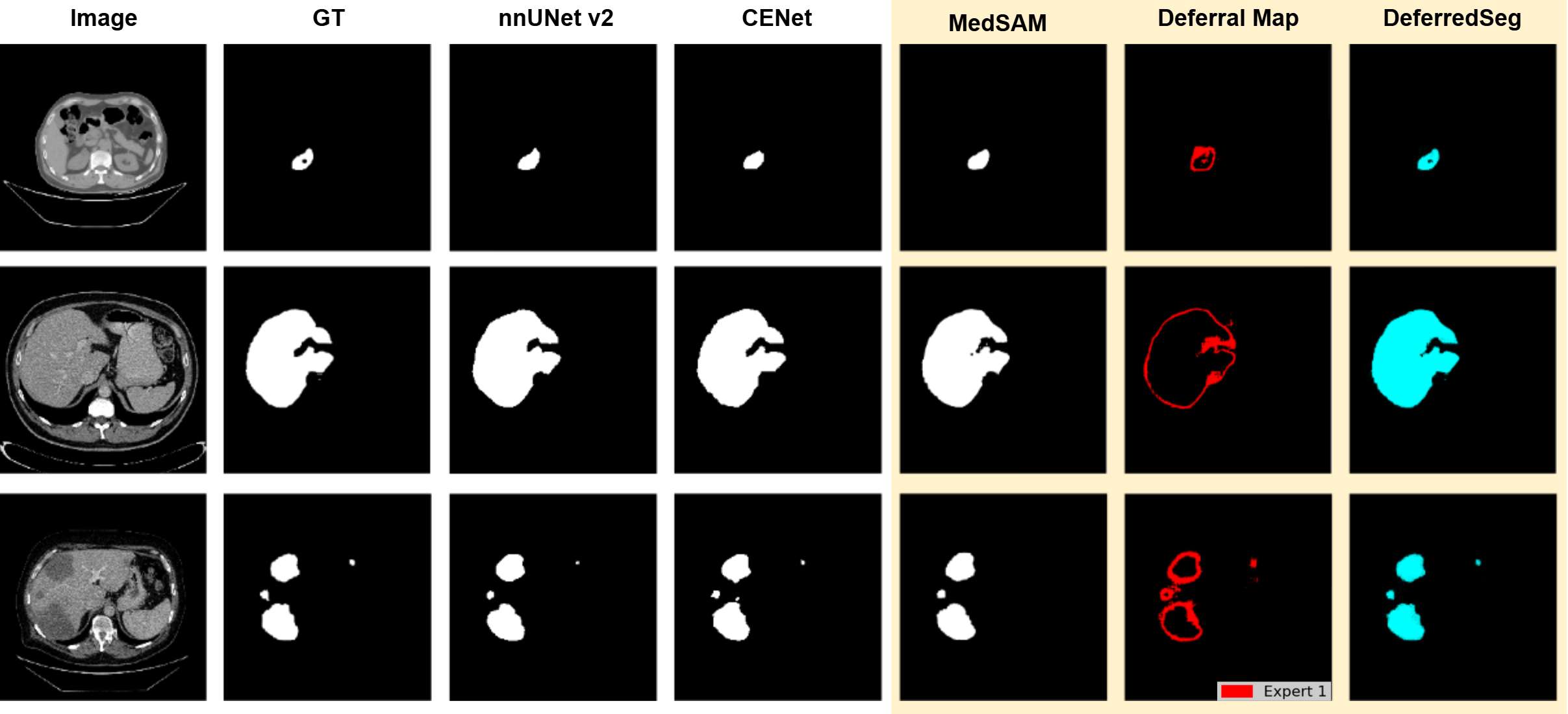}\\[-1pt]
    \includegraphics[width=0.98\linewidth]{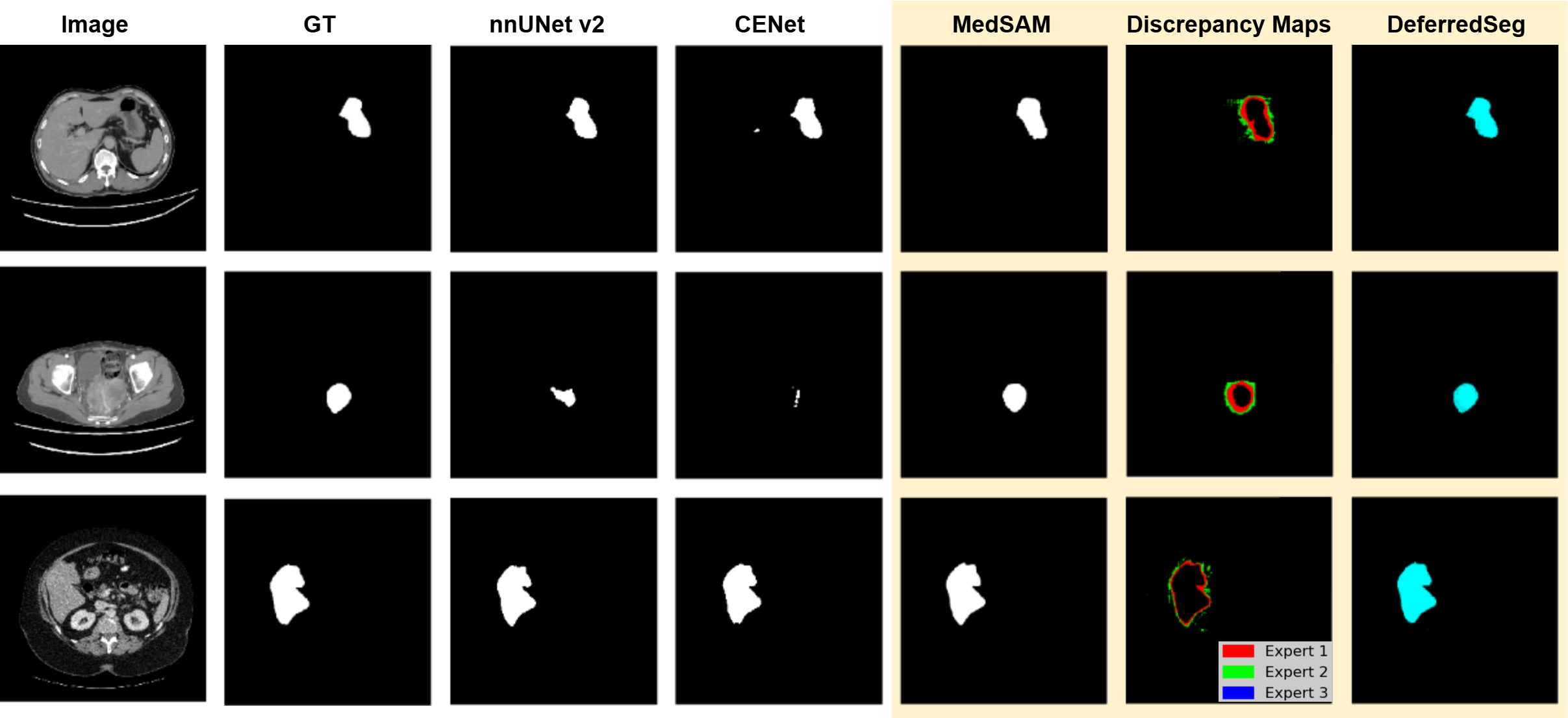}\\[-1pt]
    {\small (a)\phantomsubcaption\label{fig:seg_results_a}}
  \end{minipage}
  \hfill
  \begin{minipage}[t]{0.23\linewidth}
  \vspace{0pt}
    \centering
    \includegraphics[width=\linewidth]{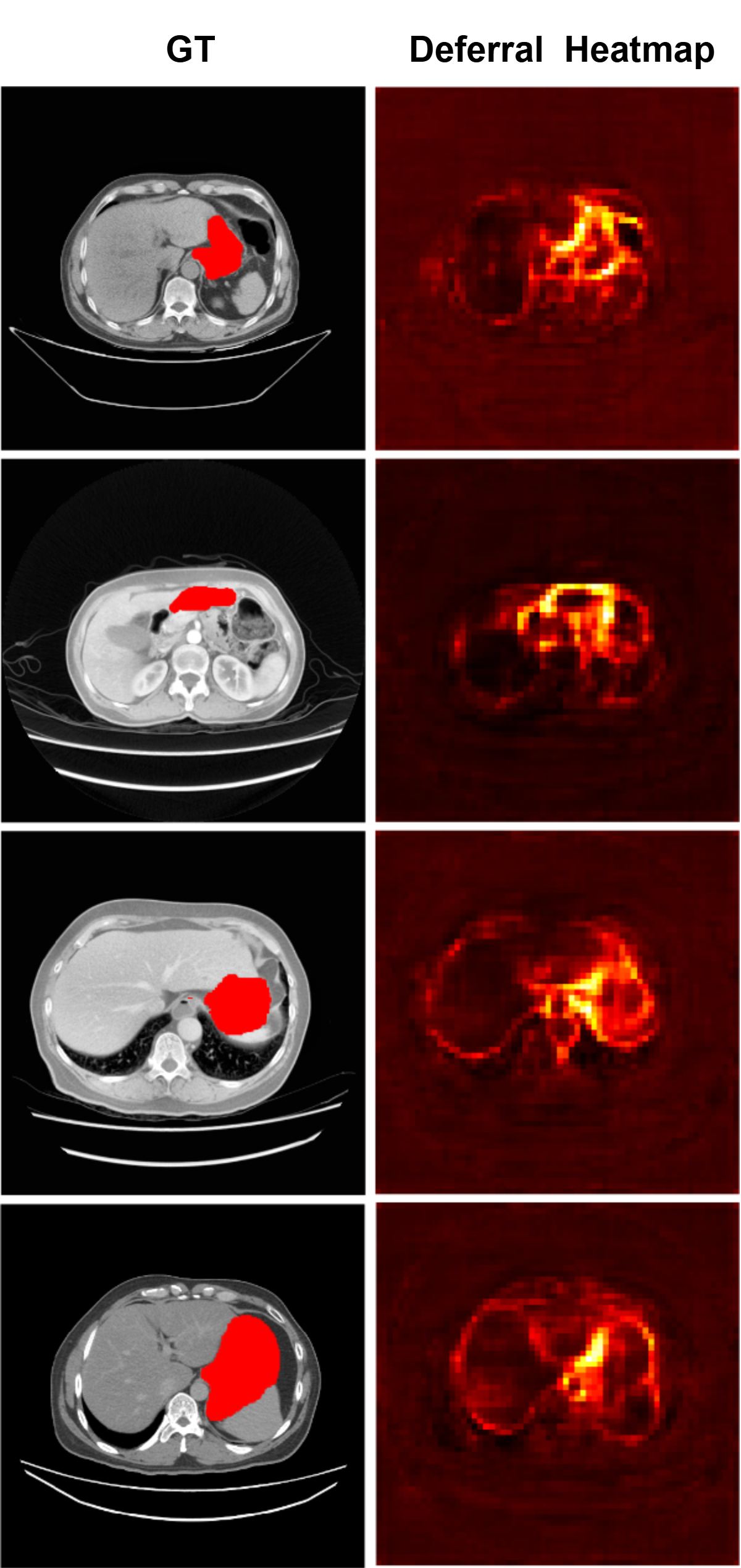}\\[-1pt]
    {\small (b)\phantomsubcaption\label{fig:seg_results_b}}

    \vspace{0pt}
    \includegraphics[width=\linewidth]{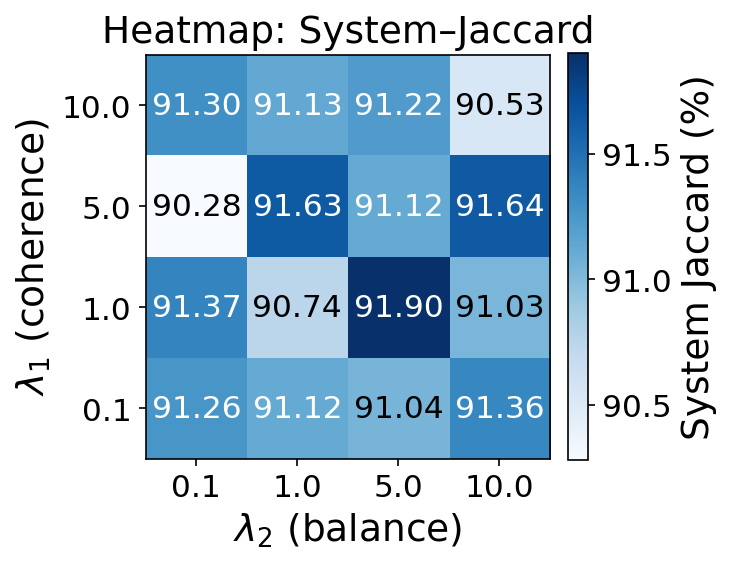}\\[-1pt]
    {\small (c)\phantomsubcaption\label{fig:seg_results_c}}
  \end{minipage}

  \vspace{-2pt}
  \caption{
     (a) Qualitative results under the \emph{1-expert} and \emph{3-expert} settings.
     (b) Deferral heatmap, where brighter regions indicate higher expert-assignment probability.
     (c) System Jaccard heatmap over coherence weight $\lambda_1$ and balance weight $\lambda_2$, with the best performance obtained at $\lambda_1{=}1$ and $\lambda_2{=}5$.
  }
  \label{fig:seg_results_all}
\end{figure}

\textbf{Expert Routing and System Predictions.} 
We compare per-slice results for the one- and three-expert configurations
against MedSAM, nnU-Net v2, and CENet. As shown in
Fig.~\ref{fig:seg_results_a}, $\nm$ produces sharper contours and fewer
false positives by deferring ambiguous pixels, particularly near organ
boundaries and in small, low-contrast regions. In the one-expert setting,
routing focuses on the most difficult pixels, whereas in the multi-expert
setting it remains boundary-concentrated but becomes more fine-grained,
allowing different experts to handle different types of difficult boundary
pixels. Thus, deferral is selective rather than uniformly distributed
across the image. Additional qualitative results are provided in
Appendix~\ref{sec:appendix_additional_visualization}.

\textbf{Deferral Heatmap Visualization.}
To inspect where deferral tends to occur, we visualize the \emph{deferral heatmap} together with the ground-truth mask in Fig.~\ref{fig:seg_results_b}. The heatmap represents the continuous aggregated expert-assignment probability $\tilde{\mathcal{M}}$, with brighter regions indicating a stronger tendency to defer pixels to expert branches, typically around ambiguous boundaries or low-contrast regions. In contrast, the final routed expert pixels are obtained by applying the argmax rule in Eq.~\eqref{eq:routing_decision} to the routing logits.

\subsection{Ablation Study and Analysis}\label{sec:ablation}
\subsubsection{Selection of Hyperparameters $(\lambda_1,\lambda_2)$}
A grid search is conducted on PROMISE12 over $\lambda_1,\lambda_2 \!\in\! \{0.1,1,5,10\}$, resulting in 16 configurations.
For brevity, the main text visualises only the System performance heatmap based on the Jaccard index (Fig.~\ref{fig:seg_results_c}).
The complete numerical results covering all three branches (System / Expert / Model) and metrics (Jaccard / DSC / Sensitivity) are reported in~\ref{app:coherence_balance} (Tab.~\ref{tab:comparison_lambdas}), together with the full set of nine heatmaps (Fig.~\ref{fig:lambda_heatmap_9}).
Across all configurations, the performance consistently peaks at $\lambda_1{=}1$ and $\lambda_2{=}5$, which are therefore adopted in the main experiments. 
To further assess whether this hyper-parameter choice is dataset-specific, we additionally conduct a sensitivity analysis on LiTS-Liver in~\ref{app:hyperparameter_more_datasets}.
\subsubsection{Effect of Loss Components}  
We conduct ablation experiments on PROMISE12 to examine the contribution of each loss component under both single-expert ($J{=}1$) and multi-expert ($J{=}3$) settings. All ablation configurations are independently trained with three random seeds. Results are summarised in Tab.~\ref{tab:ablation}.

\textbf{Single-expert.}
The results show that adding the SC Loss consistently improves the overall segmentation performance, indicating that explicit supervision on the deferral map remains beneficial in the single-expert setting.

\textbf{Multi-expert.}
With three experts and only the DC Loss, the system obtains 95.39$\pm$1.64\% DSC, which is lower than the single-expert case, indicating that simply increasing the number of experts can dilute routing. Adding the SC Loss brings small but consistent improvements, while the LB Penalty yields a larger gain. Combining SC and LB produces the best overall results, confirming that spatial coherence and balanced expert utilization are both important for effective multi-expert collaboration.

\begin{figure}[t]
  \centering
  \begin{minipage}[c]{0.42\textwidth}
    \centering
    \setlength{\tabcolsep}{1pt} %
    \renewcommand{\arraystretch}{1.05}
    \footnotesize %
    \captionof{table}{System performance with increasing numbers of complementary experts on PROMISE12. Values are averaged over three random seeds.}
    \label{tab:complementary_summary}
    \begin{tabular}{c|l|ccc}
      \toprule
      $J$ & Experts & Jaccard & DSC & Sens. \\
      \midrule
      1 & E1 & 91.04 & 95.24 & 95.54 \\
      2 & E1, E6 & 89.03 & 94.10 & 93.58 \\
      3 & E1, E2, E3 & 92.45 & 95.99 & 96.38 \\
      4 & E1, E2, E3, E4 & 94.06 & 96.83 & 97.43 \\
      5 & E1, E2, E3, E4, E5 & \textbf{94.40} & \textbf{96.88} & \textbf{97.46} \\
      \bottomrule
    \end{tabular}
  \end{minipage}%
  \hfill
  \begin{minipage}[c]{0.55\textwidth}
    \centering
    \includegraphics[width=\linewidth]{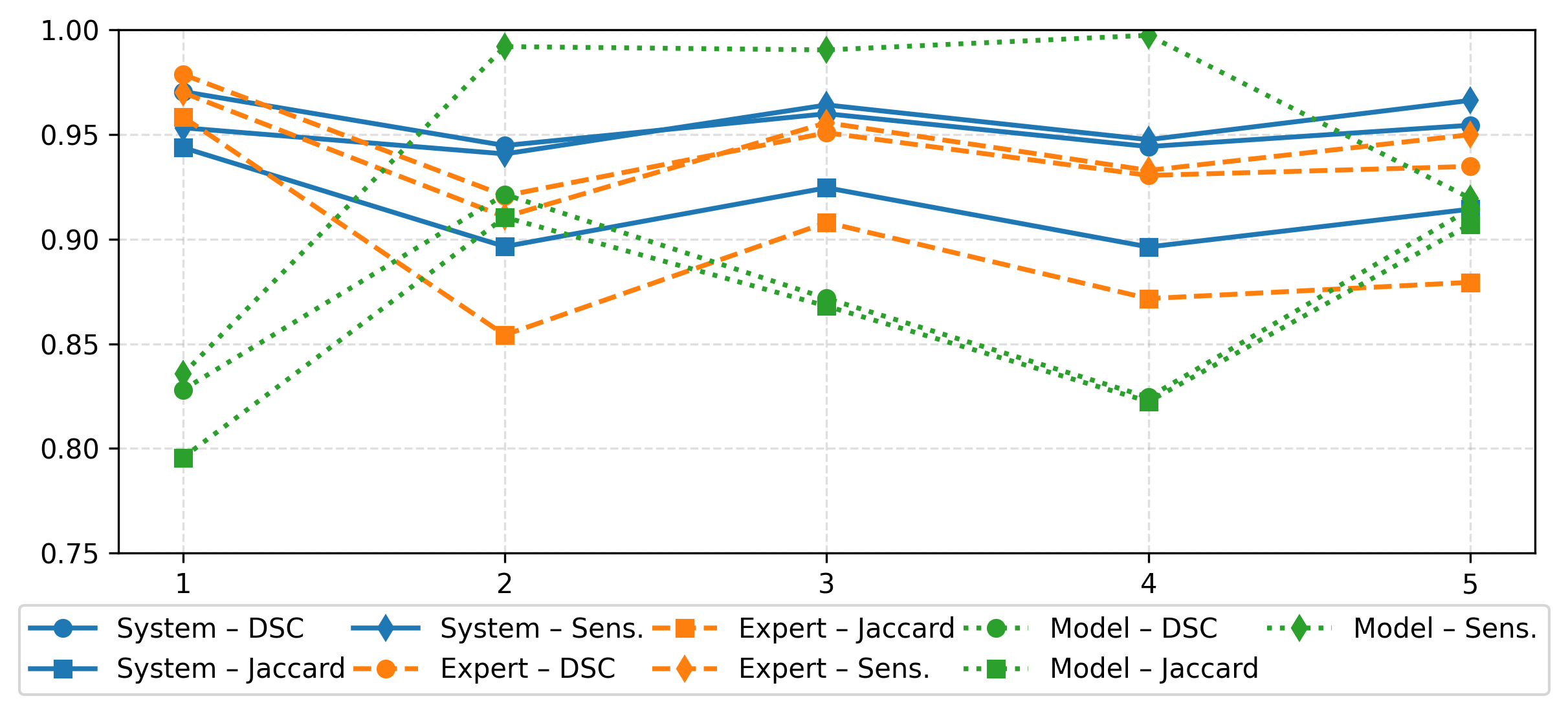}
    \caption{Expert scalability. System DSC and Jaccard peak at $J{=}1$, dip at $J{=}2$, and level off for $J\ge 3$. Sensitivity peaks at $J{=}5$.}
    \label{fig:expert_scaling}
  \end{minipage}
\end{figure}


\subsubsection{Expert Scalability Analysis}\label{sec:scalability}

To assess scalability, we evaluate $\nm$ on PROMISE12 with $J\!\in\!\{1,2,3,4,5\}$ synthetic experts under identical training settings.
The expert sets are drawn from a fixed pool with comparable FG/BG accuracy (see ~\ref{app:scalability_table}, Tab.~\ref{tab:scalability_expert_settings}), so performance differences primarily reflect how well the deferral mechanism fuses multiple experts.

From the quantitative results in \ref{app:scalability_table} (Tab.~\ref{tab:num_experts_horizontal}), the System obtains the highest overlap with a single expert ($J{=}1$). As the number of experts increases, DSC/Jaccard generally decrease or fluctuate, while Sensitivity improves and peaks at $J{=}5$. 

Overall, Fig.~\ref{fig:expert_scaling} and Tab.~\ref{tab:num_experts_horizontal} jointly reveal a clear precision--recall trade-off: adding more experts improves Sensitivity but slightly reduces overlap due to increased routing dispersion. Nevertheless, the System maintains strong DSC/Jaccard even at $J{=}5$, indicating that our pixel-wise deferral mechanism scales robustly with the number of experts on the MedSAM backbone.

\textbf{Complementary Experts.}
We further study scalability under \emph{complementary expert} configurations. 
Seven synthetic experts with distinct accuracy profiles on FG, BG, and BD regions are defined (\ref{sec:appendix_complementary_experts}, Tab.~\ref{tab:complementary_experts}); by selecting different subsets of these experts, we emulate varying levels of complementarity and specialisation.

As shown in Tab.~\ref{tab:complementary_summary}, the benefit of multiple experts is not strictly monotonic. The two-expert setting performs worse than the single-expert setting, suggesting that heterogeneous experts may introduce routing ambiguity when their strengths are not sufficiently complementary. However, with three or more complementary experts, System performance recovers and further improves, reaching the best Jaccard, DSC, and Sensitivity at $J{=}5$. These results suggest that multi-expert deferral benefits from complementary regional strengths rather than simply from including more experts. The full System / Expert / Model results with standard deviations are reported in~\ref{sec:appendix_complementary_experts} (Tab.~\ref{tab:complementary_num_results}).

\section{Conclusion}

We propose $\nm$, a pixel-wise deferral framework for medical image segmentation. By combining routing channels with deferral-aware, spatial-coherence, and load-balancing objectives, $\nm$ provides a pixel-wise mechanism for Human--AI collaboration between a base segmentor and one or more expert branches. Experiments with synthetic expert masks and clinician annotations show that $\nm$ improves System-level segmentation performance across multiple datasets, particularly in small, low-contrast, and ambiguous regions.

The main strengths of $\nm$ are its pixel-wise deferral mechanism, its support for multi-expert collaboration, and its compatibility with strong segmentation backbones such as MedSAM. 
Importantly, $\nm$ selectively routes only a small subset of pixels to expert branches during inference, as quantified by the pixel-level deferral ratio. In an intended prospective workflow, experts could focus on the difficult regions assigned by the deferral function rather than inspecting the entire image, although the actual efficiency of this workflow remains to be evaluated.

Several aspects remain worth further investigation. First, although $\nm$ reduces the proportion of pixels assigned to expert branches during inference, its training stage still relies on previously annotated expert data. Therefore, the availability and annotation cost of high-quality expert labels remain practical factors to consider. Second, the effectiveness of $\nm$ may be influenced by the reliability, regional expertise, and complementarity of the available experts. Third, our current real-expert evaluation uses pre-existing clinician annotations from Chaksu, QUBIQ, and CURVAS in an offline setting. Broader prospective studies across more organs, imaging modalities, institutions, and clinical tasks are still needed to assess clinical generality and interaction efficiency. Finally, practical deployment will require careful integration of deferred-region expert annotation into clinical workflows, including efficient interaction interfaces, manageable review time, compatibility with existing clinical systems.

\section*{Declaration of Generative AI and AI-assisted technologies in the writing process}
During the preparation of this work, the authors used generative AI and AI-assisted technologies for grammar, formatting, and spelling checks. The authors reviewed and edited the manuscript and take full responsibility for its content.

\bibliographystyle{elsarticle-num}
\bibliography{references}

\appendix
\newpage

\section{Interactive Expert Annotation Interface}
\label{app:ui}

\begin{figure}
  \centering
  \includegraphics[width=\textwidth]{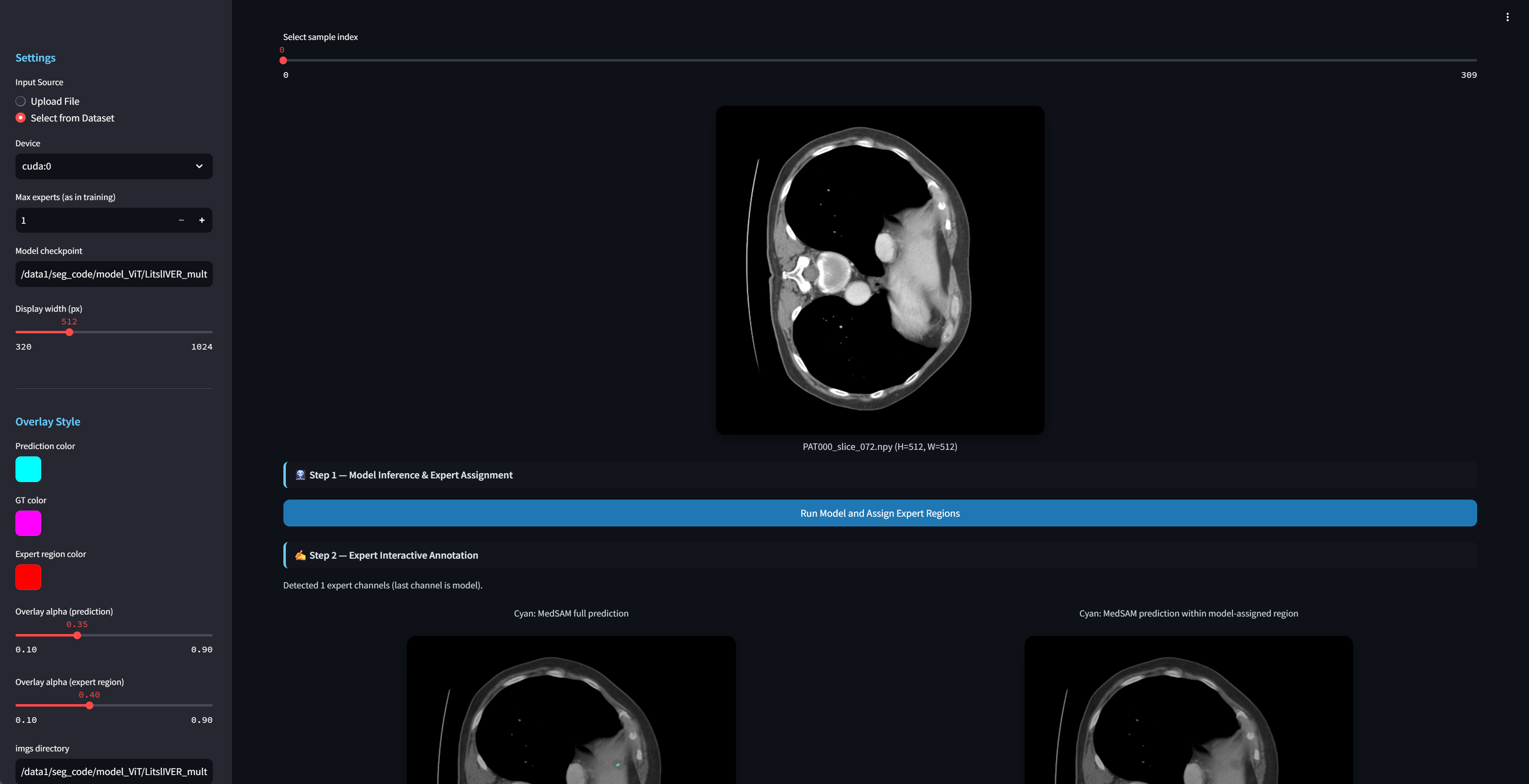}
  \caption{$\nm$ interactive expert-annotation interface. 
  The page shows (top) model prediction and routing results, (middle) per-expert canvases for interactive corrections, and (bottom) the fused segmentation and ground-truth comparison.}
  \label{fig:ui_deferredseg_streamlit}
\end{figure}

To demonstrate the intended inference-time Human--AI collaboration
workflow, we implement a proof-of-concept Streamlit interface that allows
experts to provide local corrections within regions deferred by $\nm$.
The interface demonstrates the technical workflow of expert-region
assignment, local correction, and final System fusion. However, its
usability, annotation time, interaction cost, and clinical efficiency
remain to be evaluated through prospective clinician studies. The interface is organised into three stages (see Fig.~\ref{fig:ui_deferredseg_streamlit}):

\begin{itemize}
  \item \textbf{Step 1 -- Model inference \& expert assignment.}
  The user uploads a slice (or selects one from the built-in dataset browser). $\nm$ runs the MedSAM-based segmentor and routing function to obtain the base prediction and pixel-wise routing scores, and automatically partitions the image into $J$ expert regions plus a model-only region.

  \item \textbf{Step 2 -- Expert interactive annotation.}
  For each expert region, the interface creates a dedicated canvas that only highlights pixels assigned to that expert. 
  The human expert draws corrections \emph{within their own assigned region} using freehand tools, providing refined labels exactly where their expertise is requested, while non-assigned areas remain untouched in that canvas.

  \item \textbf{Step 3 -- Fusion \& comparison.}
  The corrected expert masks are fused with the model branch via the same deferral rule as in the main framework, producing a final segmentation mask. 
  The interface visualises the fused result and, when available, overlays the ground-truth mask for qualitative comparison.
\end{itemize}

A screen-capture video demonstrating the full workflow of this interface (including loading data, assigning expert regions, annotating, and fusing the final result) is provided in the supplementary material.

\section{Dataset Details}\label{app:datasets_details}
\textbf{PROMISE12}
The PROMISE12 challenge provides 50 T2-weighted MRI volumes for training and 30 hidden test cases, each annotated with a whole-prostate mask%
\footnote{\url{https://promise12.grand-challenge.org/}}.
We resample all volumes to \(1\,\text{mm}^3\) isotropic voxel spacing. This dataset targets single-organ segmentation in MRI and serves as a testbed for assessing deferral sensitivity near ambiguous prostate boundaries.

\textbf{LiTS}
The LiTS challenge offers 201 abdominal CT scans, with 131 publicly labeled volumes containing both liver and tumour masks, and 70 held-out cases for blind testing%
\footnote{\url{https://kaiko-ai.github.io/eva/main/datasets/lits/}}.
We clip CT intensities to \([-250,250]\) Hounsfield units (HU) and linearly
scale the clipped values to the \([0,1]\) range. This dataset tests the model's ability to balance large-organ and small-lesion segmentation under significant scale disparity.

\textbf{AMOS22}
AMOS22 contains 500 CT and 100 MRI scans, each annotated for 15 abdominal organs%
\footnote{\url{https://amos22.grand-challenge.org/}}.
We follow the official split, resample all volumes to \(1\,\text{mm}\) isotropic spacing, and clip CT intensities to \([-250,250]\) HU, and normalise MRI volumes via whole-volume min–max scaling. This multi-organ dataset provides a comprehensive setting for evaluating scalability and expert routing in complex anatomical contexts.
\section{Implementation Details}
\label{app:implementation_details}

\subsubsection{Training}
The MedSAM encoder and prompt modules are frozen, and only the reconfigured decoder, routing branches, and deferral predictor are trained using AdamW with a learning rate of \(1\times10^{-4}\), StepLR scheduling, an effective batch size of 8, early stopping, and mixed precision. For all trainable methods reported with mean and standard deviation, we repeat the experiments over three random seeds and report the mean and standard deviation.
\section{Complete Quantitative Results}
\label{sec:appendix_sota_summary}

As a complement to Tab.~\ref{tab:sota_summary} in the main text, Tab.~\ref{tab:sota_summary_all} reports the complete quantitative results across multiple datasets, organs, and evaluation metrics, including standard deviations as well as the Expert and Model branches.

\begin{table}[!htbp]
\enlargethispage{\baselineskip}
  \centering
 \caption{Comparison with fully automatic reference baselines and the expert-assisted PatchCorr baseline across multiple datasets, organs, and evaluation metrics (\%). Row-wise best values are \textbf{bold}.}
  \label{tab:sota_summary_all}
  \setlength{\tabcolsep}{1.3pt}
  \renewcommand{\arraystretch}{0.85}
  \scriptsize
  \resizebox{\textwidth}{!}{%
  \begin{tabular}{llccccc ccc ccc}
    \toprule
    \textbf{Dataset} & \textbf{Metric} 
      & \textbf{nnUNet v2} & \textbf{CENet} & \textbf{MedSAM} & \textbf{MedSAM}
      & \textbf{PatchCorr}
      & \multicolumn{3}{c|}{\textbf{Ours (1 expert)}}
      & \multicolumn{3}{c}{\textbf{Ours (3 experts)}} \\
    \cmidrule(lr){8-10} \cmidrule(lr){11-13}
    & 
      & \tiny{(MICCAI 2024)} & \tiny{(MICCAI 2025)} & \tiny{(NC 2024)} & \tiny{(Fine-tuned)}
      & \tiny{(MIDL 2024)}
      & System & Expert & Model
      & System & Expert & Model \\    
    \midrule
    \multirow{3}{*}{\shortstack{Prostate\\(PROMISE12)}}
      & DSC     
        & 90.49 $\pm$ 0.12 & 68.96 $\pm$ 0.16 & 84.16  & 86.17 $\pm$ 1.53
        & 92.24 $\pm$ 0.03
        & \textbf{97.06 $\pm$ 0.05} & \textcolor{gray}{97.88 $\pm$ 0.01} & \textcolor{gray}{82.80 $\pm$ 2.75}
        & 96.00 $\pm$ 0.14 & \textcolor{gray}{95.09 $\pm$ 0.19} & \textcolor{gray}{87.17 $\pm$ 1.99} \\        
      & Jaccard 
        & 82.70 $\pm$ 0.20 & 54.62 $\pm$ 0.13 & 73.01 & 76.08 $\pm$ 2.32
        & 85.98 $\pm$ 0.05
        & \textbf{94.38 $\pm$ 0.08} & \textcolor{gray}{95.85 $\pm$ 0.01} & \textcolor{gray}{79.56 $\pm$ 2.68}
        & 92.46 $\pm$ 0.24 & \textcolor{gray}{90.80 $\pm$ 0.32} & \textcolor{gray}{86.80 $\pm$ 1.91} \\     
      & Sens.   
        & 91.78 $\pm$ 0.53 & 75.20 $\pm$ 0.10 & 76.70 & 84.97 $\pm$ 6.97
        & 94.45 $\pm$ 0.04
        & 95.33 $\pm$ 0.09 & \textcolor{gray}{97.01 $\pm$ 0.00} & \textcolor{gray}{83.55 $\pm$ 1.37}
        & \textbf{96.42 $\pm$ 0.04} & \textcolor{gray}{95.57 $\pm$ 0.04} & \textcolor{gray}{99.05 $\pm$ 0.64} \\

    \midrule
 \multirow{3}{*}{\shortstack{Liver\\(LiTS)}}
      & DSC     
        & 96.62 $\pm$ 0.05 & 93.61 $\pm$ 0.10 & 91.64 & 91.69 $\pm$ 0.06
        & 92.86 $\pm$ 0.00
        & \textbf{97.96 $\pm$ 0.15} & \textcolor{gray}{95.38 $\pm$ 0.09} & \textcolor{gray}{95.07 $\pm$ 0.14}
        & 96.15 $\pm$ 0.51 & \textcolor{gray}{97.17 $\pm$ 0.31} & \textcolor{gray}{93.69 $\pm$ 0.98} \\
      & Jaccard 
        & 93.48 $\pm$ 0.10 & 89.52 $\pm$ 0.07 & 86.43 & 86.49 $\pm$ 0.07
        & 88.83 $\pm$ 0.01
        & \textbf{96.30 $\pm$ 0.16} & \textcolor{gray}{91.34 $\pm$ 0.07} & \textcolor{gray}{94.13 $\pm$ 0.13}
        & 93.70 $\pm$ 0.64 & \textcolor{gray}{95.16 $\pm$ 0.35} & \textcolor{gray}{91.01 $\pm$ 1.04} \\
      & Sens.   
        & \textbf{98.37 $\pm$ 0.11} & 93.85 $\pm$ 0.14 & 93.98 & 94.06 $\pm$ 0.04
        & 98.08 $\pm$ 0.01
        & 96.83 $\pm$ 0.23 & \textcolor{gray}{92.02 $\pm$ 0.01} & \textcolor{gray}{95.57 $\pm$ 0.60}
        & 95.47 $\pm$ 0.60 & \textcolor{gray}{97.03 $\pm$ 0.03} & \textcolor{gray}{93.16 $\pm$ 0.65} \\
    \midrule
    \multirow{3}{*}{\shortstack{Tumor\\(LiTS)}}
      & DSC     
        & 68.04 $\pm$ 1.88 & 75.35 $\pm$ 0.52 & 59.62 & 59.20 $\pm$ 0.26
        & 65.42 $\pm$ 0.05
        & \textbf{91.01 $\pm$ 4.30} & \textcolor{gray}{93.88 $\pm$ 3.16} & \textcolor{gray}{67.34 $\pm$ 25.33}
        & 81.82 $\pm$ 12.90 & \textcolor{gray}{83.51 $\pm$ 9.61} & \textcolor{gray}{68.52 $\pm$ 5.86} \\
      & Jaccard 
        & 52.72 $\pm$ 1.10 & 64.91 $\pm$ 0.36 & 47.83 & 47.55 $\pm$ 0.14
        & 54.91 $\pm$ 0.06
        & \textbf{85.81 $\pm$ 6.60} & \textcolor{gray}{89.64 $\pm$ 5.36} & \textcolor{gray}{64.03 $\pm$ 26.14}
        & 74.59 $\pm$ 15.04 & \textcolor{gray}{78.52 $\pm$ 6.72} & \textcolor{gray}{63.42 $\pm$ 11.30} \\
      & Sens.   
        & 84.99 $\pm$ 0.52 & 76.38 $\pm$ 0.65 & 76.18 & 75.97 $\pm$ 0.62
        & 87.75 $\pm$ 0.10
        & \textbf{88.99 $\pm$ 6.67} & \textcolor{gray}{93.81 $\pm$ 2.80} & \textcolor{gray}{73.00 $\pm$ 24.24}
        & 84.39 $\pm$ 16.40 & \textcolor{gray}{95.65 $\pm$ 3.31} & \textcolor{gray}{80.04 $\pm$ 16.58} \\
   \midrule
   \multirow{3}{*}{Spleen}
      & DSC     
        & 94.29 $\pm$ 0.07 & 89.62 $\pm$ 0.25 & 89.21 & 94.57 $\pm$ 0.39
        & 93.80 $\pm$ 0.00
        & 97.75 $\pm$ 0.46 & \textcolor{gray}{95.52 $\pm$ 2.59} & \textcolor{gray}{97.69 $\pm$ 0.06}
        & \textbf{98.14 $\pm$ 0.49} & \textcolor{gray}{95.74 $\pm$ 2.47} & \textcolor{gray}{98.28 $\pm$ 0.11} \\
      & Jaccard 
        & 90.83 $\pm$ 0.11 & 84.79 $\pm$ 0.26 & 81.74 & 90.77 $\pm$ 0.50
        & 89.33 $\pm$ 0.00
        & 95.86 $\pm$ 0.85 & \textcolor{gray}{91.64 $\pm$ 4.74} & \textcolor{gray}{96.52 $\pm$ 0.06}
        & \textbf{96.59 $\pm$ 0.93} & \textcolor{gray}{92.06 $\pm$ 4.57} & \textcolor{gray}{97.59 $\pm$ 0.11} \\
      & Sens.   
        & 95.21 $\pm$ 0.53 & 89.41 $\pm$ 0.28 & 90.78 & 95.02 $\pm$ 0.37
        & 95.41 $\pm$ 0.00
        & 98.06 $\pm$ 0.87 & \textcolor{gray}{95.50 $\pm$ 4.95} & \textcolor{gray}{99.00 $\pm$ 0.01}
        & \textbf{98.34 $\pm$ 0.94} & \textcolor{gray}{95.34 $\pm$ 4.72} & \textcolor{gray}{99.61 $\pm$ 0.01} \\
    \midrule
    \multirow{3}{*}{Right Kidney}
      & DSC     
        & 90.26 $\pm$ 0.74 & 91.34 $\pm$ 0.86 & 89.39 & 94.83 $\pm$ 0.07
        & 93.76 $\pm$ 0.00
        & \textbf{98.34 $\pm$ 0.68} & \textcolor{gray}{95.88 $\pm$ 2.60} & \textcolor{gray}{98.22 $\pm$ 0.02}
        & \textbf{98.34 $\pm$ 0.51} & \textcolor{gray}{96.08 $\pm$ 2.56} & \textcolor{gray}{98.30 $\pm$ 0.00} \\
      & Jaccard 
        & 85.99 $\pm$ 0.72 & 86.47 $\pm$ 0.93 & 81.98 & 90.85 $\pm$ 0.08
        & 89.05 $\pm$ 0.01
        & \textbf{96.85 $\pm$ 1.30} & \textcolor{gray}{92.23 $\pm$ 4.80} & \textcolor{gray}{97.55 $\pm$ 0.04}
        & 96.84 $\pm$ 0.97 & \textcolor{gray}{92.61 $\pm$ 4.74} & \textcolor{gray}{97.66 $\pm$ 0.01} \\
      & Sens.   
        & 85.86 $\pm$ 0.51 & 91.27 $\pm$ 1.16 & 91.70 & 95.25 $\pm$ 0.04
        & 96.46 $\pm$ 0.00
        & \textbf{98.69 $\pm$ 1.31} & \textcolor{gray}{95.51 $\pm$ 4.94} & \textcolor{gray}{99.67 $\pm$ 0.01}
        & 98.49 $\pm$ 0.96 & \textcolor{gray}{95.42 $\pm$ 4.84} & \textcolor{gray}{99.80 $\pm$ 0.02} \\
    \midrule
    \multirow{3}{*}{Left kidney}
      & DSC     
        & 93.67 $\pm$ 0.29 & 91.61 $\pm$ 1.36 & 89.63 & 94.74 $\pm$ 0.09
        & 93.42 $\pm$ 0.00
        & 98.14 $\pm$ 0.63 & \textcolor{gray}{95.68 $\pm$ 2.60} & \textcolor{gray}{97.91 $\pm$ 0.04}
        & \textbf{98.16 $\pm$ 0.56} & \textcolor{gray}{96.12 $\pm$ 2.51} & \textcolor{gray}{98.34 $\pm$ 0.01} \\        
      & Jaccard 
        & 88.99 $\pm$ 0.37 & 86.92 $\pm$ 1.48 & 82.33 & 90.86 $\pm$ 0.12
        & 88.63 $\pm$ 0.00
        & 96.60 $\pm$ 1.18 & \textcolor{gray}{91.94 $\pm$ 4.78} & \textcolor{gray}{97.13 $\pm$ 0.04}
        & \textbf{96.63 $\pm$ 1.09} & \textcolor{gray}{92.72 $\pm$ 4.68} & \textcolor{gray}{97.85 $\pm$ 0.01} \\        
      & Sens.   
        & 93.21 $\pm$ 1.31 & 91.39 $\pm$ 1.81 & 93.11 & 94.67 $\pm$ 0.29
        & 97.62 $\pm$ 0.00
        & \textbf{98.72 $\pm$ 1.23} & \textcolor{gray}{95.50 $\pm$ 4.95} & \textcolor{gray}{99.53 $\pm$ 0.04}
        & 98.25 $\pm$ 1.12 & \textcolor{gray}{95.42 $\pm$ 4.84} & \textcolor{gray}{99.74 $\pm$ 0.07} \\
    \midrule
    \multirow{3}{*}{Gallbladder}
      & DSC     
        & 76.73 $\pm$ 0.21 & 62.58 $\pm$ 3.72 & 79.60 & 86.69 $\pm$ 0.67
        & 90.82 $\pm$ 0.00
        & \textbf{96.57 $\pm$ 1.42} & \textcolor{gray}{95.19 $\pm$ 2.56} & \textcolor{gray}{90.84 $\pm$ 0.25}
        & 95.27 $\pm$ 1.17 & \textcolor{gray}{93.75 $\pm$ 2.55} & \textcolor{gray}{92.22 $\pm$ 0.67} \\
      & Jaccard 
        & 68.31 $\pm$ 0.28 & 45.66 $\pm$ 3.39 & 68.58 & 78.77 $\pm$ 0.79
        & 84.62 $\pm$ 0.00
        & \textbf{93.88 $\pm$ 2.65} & \textcolor{gray}{91.29 $\pm$ 4.70} & \textcolor{gray}{90.05 $\pm$ 0.22}
        & 91.82 $\pm$ 2.17 & \textcolor{gray}{89.19 $\pm$ 4.59} & \textcolor{gray}{91.40 $\pm$ 0.70} \\
      & Sens.   
        & 81.95 $\pm$ 0.52 & 50.48 $\pm$ 4.23 & 79.56 & 88.46 $\pm$ 1.66
        & 92.20 $\pm$ 0.01
        & \textbf{97.27 $\pm$ 2.90} & \textcolor{gray}{95.48 $\pm$ 5.00} & \textcolor{gray}{99.19 $\pm$ 0.01}
        & 96.62 $\pm$ 2.24 & \textcolor{gray}{95.40 $\pm$ 4.78} & \textcolor{gray}{99.56 $\pm$ 0.17} \\
    \midrule
    \multirow{3}{*}{Esophagus}
      & DSC     
        & 82.71 $\pm$ 0.05 & 66.44 $\pm$ 0.54 & 65.72 & 79.67 $\pm$ 0.51
        & 90.96 $\pm$ 0.01
        & 96.14 $\pm$ 1.81 & \textcolor{gray}{95.62 $\pm$ 2.59} & \textcolor{gray}{89.05 $\pm$ 0.31}
        & \textbf{96.44 $\pm$ 1.78} & \textcolor{gray}{95.82 $\pm$ 2.48} & \textcolor{gray}{89.44 $\pm$ 0.01} \\   
      & Jaccard 
        & 72.07 $\pm$ 0.05 & 55.24 $\pm$ 0.59 & 51.50 & 68.37 $\pm$ 0.09
        & 84.10 $\pm$ 0.01
        & 92.78 $\pm$ 3.36 & \textcolor{gray}{91.76 $\pm$ 4.78} & \textcolor{gray}{87.19 $\pm$ 0.30}
        & \textbf{93.28 $\pm$ 3.33} & \textcolor{gray}{92.14 $\pm$ 4.60} & \textcolor{gray}{87.96 $\pm$ 0.01} \\
      & Sens.   
        & 84.43 $\pm$ 0.15 & 65.63 $\pm$ 0.73 & 67.36 & 78.58 $\pm$ 0.31
        & 88.46 $\pm$ 0.01
        & \textbf{96.50 $\pm$ 3.48} & \textcolor{gray}{95.50 $\pm$ 4.95} & \textcolor{gray}{96.98 $\pm$ 0.17}
        & 96.43 $\pm$ 3.46 & \textcolor{gray}{95.36 $\pm$ 4.75} & \textcolor{gray}{98.25 $\pm$ 0.16} \\
    \midrule
      \multirow{3}{*}{\shortstack{Liver\\(AMOS22)}}
      & DSC     
        & 97.17 $\pm$ 0.05 & 90.90 $\pm$ 0.43 & 86.78 & 95.15 $\pm$ 0.52
        & 89.29 $\pm$ 0.00
        & 97.54 $\pm$ 0.49 & \textcolor{gray}{95.40 $\pm$ 2.56} & \textcolor{gray}{96.93 $\pm$ 0.02}
        & \textbf{97.87 $\pm$ 0.45} & \textcolor{gray}{95.65 $\pm$ 2.47} & \textcolor{gray}{97.15 $\pm$ 0.07} \\    
      & Jaccard 
        & 94.57 $\pm$ 0.09 & 86.80 $\pm$ 0.44 & 78.67 & 91.82 $\pm$ 0.11
        & 82.85 $\pm$ 0.00
        & 95.48 $\pm$ 0.93 & \textcolor{gray}{91.42 $\pm$ 4.71} & \textcolor{gray}{95.40 $\pm$ 0.01}
        & \textbf{96.07 $\pm$ 0.86} & \textcolor{gray}{91.85 $\pm$ 4.52} & \textcolor{gray}{95.87 $\pm$ 0.06} \\        
      & Sens.   
        & 97.37 $\pm$ 0.05 & 90.64 $\pm$ 0.73 & 91.25 & 95.44 $\pm$ 0.84
        & 95.99 $\pm$ 0.00
        & 98.44 $\pm$ 0.96 & \textcolor{gray}{95.50 $\pm$ 4.95} & \textcolor{gray}{98.92 $\pm$ 0.03}
        & \textbf{98.70 $\pm$ 0.87} & \textcolor{gray}{96.26 $\pm$ 4.70} & \textcolor{gray}{99.20 $\pm$ 0.07} \\
    \midrule
    \multirow{3}{*}{Stomach}
      & DSC     
        & 87.27 $\pm$ 0.30 & 77.07 $\pm$ 0.60 & 81.56 & 90.69 $\pm$ 0.41
        & 89.81 $\pm$ 0.00
        & \textbf{96.70 $\pm$ 0.72} & \textcolor{gray}{95.54 $\pm$ 2.56} & \textcolor{gray}{95.97 $\pm$ 0.06}
        & 96.67 $\pm$ 0.57 & \textcolor{gray}{95.62 $\pm$ 2.36} & \textcolor{gray}{95.73 $\pm$ 0.04} \\
      & Jaccard 
        & 80.14 $\pm$ 0.36 & 70.37 $\pm$ 0.50 & 70.82 & 84.98 $\pm$ 0.28
        & 83.32 $\pm$ 0.00
        & \textbf{94.03 $\pm$ 1.36} & \textcolor{gray}{91.73 $\pm$ 4.72} & \textcolor{gray}{94.27 $\pm$ 0.06}
        & 93.98 $\pm$ 1.08 & \textcolor{gray}{91.85 $\pm$ 4.36} & \textcolor{gray}{93.73 $\pm$ 0.01} \\
      & Sens.   
        & 89.78 $\pm$ 0.43 & 76.35 $\pm$ 0.81 & 81.04 & 90.46 $\pm$ 0.07
        & 92.84 $\pm$ 0.01
        & \textbf{97.62 $\pm$ 1.41} & \textcolor{gray}{95.51 $\pm$ 4.94} & \textcolor{gray}{99.02 $\pm$ 0.11}
        & \textbf{97.62 $\pm$ 1.15} & \textcolor{gray}{95.12 $\pm$ 4.49} & \textcolor{gray}{98.71 $\pm$ 0.03} \\
    \midrule
      \multirow{3}{*}{Aorta}
      & DSC     
        & 94.85 $\pm$ 0.06 & 86.35 $\pm$ 3.42 & 86.45 & 94.61 $\pm$ 0.87
        & 93.57 $\pm$ 0.00
        & 98.00 $\pm$ 0.50 & \textcolor{gray}{95.95 $\pm$ 2.57} & \textcolor{gray}{98.72 $\pm$ 0.02}
        & \textbf{98.20 $\pm$ 0.42} & \textcolor{gray}{95.97 $\pm$ 2.52} & \textcolor{gray}{98.72 $\pm$ 0.01} \\
      & Jaccard 
        & 90.40 $\pm$ 0.09 & 80.95 $\pm$ 3.51 & 78.02 & 90.36 $\pm$ 1.02
        & 89.90 $\pm$ 0.00
        & 96.34 $\pm$ 0.98 & \textcolor{gray}{92.38 $\pm$ 4.78} & \textcolor{gray}{97.96 $\pm$ 0.03}
        & \textbf{96.68 $\pm$ 0.82} & \textcolor{gray}{92.46 $\pm$ 4.68} & \textcolor{gray}{97.91 $\pm$ 0.02} \\
      & Sens.   
        & 93.75 $\pm$ 0.22 & 85.29 $\pm$ 4.15 & 87.59 & 94.35 $\pm$ 0.71
        & 94.69 $\pm$ 0.00
        & 98.11 $\pm$ 1.02 & \textcolor{gray}{95.49 $\pm$ 4.94} & \textcolor{gray}{99.33 $\pm$ 0.02}
        & \textbf{98.47 $\pm$ 0.88} & \textcolor{gray}{95.40 $\pm$ 4.82} & \textcolor{gray}{99.57 $\pm$ 0.03} \\

    \midrule
      \multirow{3}{*}{\shortstack{Inferior\\vena cava}}
      & DSC     
        & 87.84 $\pm$ 0.03 & 75.42 $\pm$ 1.07 & 77.59 & 85.42 $\pm$ 0.40
        & 92.37 $\pm$ 0.00
        & \textbf{97.02 $\pm$ 0.93} & \textcolor{gray}{95.96 $\pm$ 2.57} & \textcolor{gray}{97.30 $\pm$ 0.05}
        & 97.01 $\pm$ 0.81 & \textcolor{gray}{96.04 $\pm$ 2.48} & \textcolor{gray}{97.10 $\pm$ 0.07} \\
      & Jaccard 
        & 78.84 $\pm$ 0.07 & 66.16 $\pm$ 1.15 & 65.58 & 76.61 $\pm$ 0.19
        & 86.43 $\pm$ 0.00
        & \textbf{94.39 $\pm$ 1.78} & \textcolor{gray}{92.38 $\pm$ 4.77} & \textcolor{gray}{95.95 $\pm$ 0.04}
        & 94.38 $\pm$ 1.54 & \textcolor{gray}{92.50 $\pm$ 4.60} & \textcolor{gray}{95.47 $\pm$ 0.06} \\
      & Sens.   
        & 88.42 $\pm$ 0.17 & 85.20 $\pm$ 1.17 & 75.48 & 83.88 $\pm$ 0.62
        & 89.20 $\pm$ 0.00
        & 97.17 $\pm$ 1.82 & \textcolor{gray}{95.51 $\pm$ 4.94} & \textcolor{gray}{99.26 $\pm$ 0.02}
        & \textbf{97.23 $\pm$ 1.59} & \textcolor{gray}{95.30 $\pm$ 4.72} & \textcolor{gray}{99.00 $\pm$ 0.02} \\

    \midrule
    \multirow{3}{*}{Pancreas}
      & DSC     
        & 82.06 $\pm$ 0.23 & 62.46 $\pm$ 1.21 & 69.30 & 83.90 $\pm$ 0.09
        & 80.51 $\pm$ 0.00
        & \textbf{95.68 $\pm$ 1.62} & \textcolor{gray}{95.50 $\pm$ 2.57} & \textcolor{gray}{90.58 $\pm$ 0.56}
        & 95.39 $\pm$ 1.25 & \textcolor{gray}{95.57 $\pm$ 2.43} & \textcolor{gray}{91.02 $\pm$ 0.40} \\
      & Jaccard 
        & 71.11 $\pm$ 0.30 & 47.18 $\pm$ 1.67 & 56.06 & 74.11 $\pm$ 0.58
        & 71.06 $\pm$ 0.00
        & \textbf{92.19 $\pm$ 3.01} & \textcolor{gray}{91.72 $\pm$ 4.74} & \textcolor{gray}{87.93 $\pm$ 0.52}
        & 91.67 $\pm$ 2.32 & \textcolor{gray}{91.86 $\pm$ 4.50} & \textcolor{gray}{88.38 $\pm$ 0.43} \\
      & Sens.   
        & 85.44 $\pm$ 0.38 & 56.61 $\pm$ 0.35 & 77.23 & 86.92 $\pm$ 0.53
        & 90.46 $\pm$ 0.01
        & \textbf{96.54 $\pm$ 3.17} & \textcolor{gray}{95.51 $\pm$ 4.95} & \textcolor{gray}{96.42 $\pm$ 0.01}
        & 95.82 $\pm$ 2.41 & \textcolor{gray}{95.26 $\pm$ 4.63} & \textcolor{gray}{97.02 $\pm$ 0.04} \\
    \midrule
    \multirow{3}{*}{Duodenum}
      & DSC     
        & 74.75 $\pm$ 0.10 & 38.54 $\pm$ 9.32 & 62.14 & 79.83 $\pm$ 0.85
        & 75.44 $\pm$ 0.00
        & 94.63 $\pm$ 1.76 & \textcolor{gray}{95.44 $\pm$ 2.56} & \textcolor{gray}{86.81 $\pm$ 0.11}
        & \textbf{94.86 $\pm$ 1.56} & \textcolor{gray}{95.64 $\pm$ 2.38} & \textcolor{gray}{88.07 $\pm$ 0.09} \\    
      & Jaccard 
        & 62.07 $\pm$ 0.13 & 25.74 $\pm$ 7.95 & 48.34 & 68.50 $\pm$ 0.32
        & 65.21 $\pm$ 0.01
        & 90.38 $\pm$ 3.21 & \textcolor{gray}{91.47 $\pm$ 4.71} & \textcolor{gray}{83.03 $\pm$ 0.08}
        & \textbf{90.72 $\pm$ 2.84} & \textcolor{gray}{91.81 $\pm$ 4.40} & \textcolor{gray}{84.27 $\pm$ 0.08} \\      
      & Sens.   
        & 74.59 $\pm$ 0.15 & 30.89 $\pm$ 8.65 & 69.37 & 81.53 $\pm$ 0.62
        & 82.02 $\pm$ 0.02
        & \textbf{95.40 $\pm$ 3.37} & \textcolor{gray}{95.49 $\pm$ 4.96} & \textcolor{gray}{93.16 $\pm$ 0.09}
        & 95.38 $\pm$ 3.03 & \textcolor{gray}{95.18 $\pm$ 4.55} & \textcolor{gray}{93.58 $\pm$ 0.01} \\
    \midrule
\multirow{3}{*}{Bladder}
    & DSC     
      & 87.11 $\pm$ 0.54 & 75.38 $\pm$ 13.87 & 81.64 & 89.17 $\pm$ 0.57
      & 89.50 $\pm$ 0.00
      & \textbf{95.99 $\pm$ 1.18} & \textcolor{gray}{95.65 $\pm$ 2.58} & \textcolor{gray}{92.98 $\pm$ 0.11}
      & \textbf{95.99 $\pm$ 1.10} & \textcolor{gray}{95.46 $\pm$ 2.48} & \textcolor{gray}{92.45 $\pm$ 0.04} \\
    & Jaccard 
      & 80.74 $\pm$ 0.41 & 61.78 $\pm$ 16.36 & 71.80 & 82.72 $\pm$ 0.89
      & 83.42 $\pm$ 0.00
      & 93.27 $\pm$ 2.15 & \textcolor{gray}{91.98 $\pm$ 4.76} & \textcolor{gray}{91.42 $\pm$ 0.12}
      & \textbf{93.29 $\pm$ 2.02} & \textcolor{gray}{91.75 $\pm$ 4.54} & \textcolor{gray}{90.90 $\pm$ 0.03} \\  
    & Sens.   
      & 94.39 $\pm$ 0.35 & 69.68 $\pm$ 15.57 & 81.94 & 92.45 $\pm$ 0.26
      & 93.21 $\pm$ 0.01
      & \textbf{97.68 $\pm$ 2.26} & \textcolor{gray}{95.50 $\pm$ 4.95} & \textcolor{gray}{98.61 $\pm$ 0.31}
      & 97.62 $\pm$ 2.18 & \textcolor{gray}{95.32 $\pm$ 4.78} & \textcolor{gray}{98.62 $\pm$ 0.07} \\
    \midrule
\multirow{3}{*}{\shortstack{Prostate\\/Uterus}}
      & DSC     
        & 85.40 $\pm$ 0.23 & 41.64 $\pm$ 14.05 & 81.11 & 88.73 $\pm$ 0.15
        & 93.16 $\pm$ 0.01
        & 97.09 $\pm$ 0.98 & \textcolor{gray}{95.69 $\pm$ 1.82} & \textcolor{gray}{97.74 $\pm$ 0.03}
        & \textbf{97.19 $\pm$ 0.90} & \textcolor{gray}{95.72 $\pm$ 2.50} & \textcolor{gray}{97.32 $\pm$ 0.15} \\
      & Jaccard 
        & 75.83 $\pm$ 0.31 & 29.57 $\pm$ 12.14 & 69.65 & 80.77 $\pm$ 0.61
        & 87.77 $\pm$ 0.02
        & 94.58 $\pm$ 1.83 & \textcolor{gray}{91.90 $\pm$ 3.35} & \textcolor{gray}{96.89 $\pm$ 0.09}
        & \textbf{94.78 $\pm$ 1.72} & \textcolor{gray}{92.02 $\pm$ 4.62} & \textcolor{gray}{96.52 $\pm$ 0.15} \\
      & Sens.   
        & 88.46 $\pm$ 0.16 & 36.72 $\pm$ 15.53 & 76.65 & 91.44 $\pm$ 0.08
        & 92.42 $\pm$ 0.01
        & 97.21 $\pm$ 1.85 & \textcolor{gray}{95.50 $\pm$ 3.50} & \textcolor{gray}{99.42 $\pm$ 0.06}
        & \textbf{97.26 $\pm$ 1.80} & \textcolor{gray}{95.42 $\pm$ 4.82} & \textcolor{gray}{99.39 $\pm$ 0.06} \\
    \bottomrule
\end{tabular}}
\end{table}

\section{Complete Real-Expert Evaluation Results}
\label{sec:appendix_real_expert}

As a complement to Tab.~\ref{tab:real_expert_summary} in the main text,
Tab.~\ref{tab:real_expert_evaluation} reports the complete real-expert
evaluation results across Chaksu, QUBIQ, and CURVAS. The table includes
standard deviations as well as the System, Expert, and Model branch
results of $\nm$.

\begin{table*}[!htbp]
  \centering
  \caption{Complete real-expert evaluation across Chaksu, QUBIQ, and
  CURVAS (mean $\pm$ standard deviation, \%). Fully automatic methods are
  included as model-only references, while PatchCorr is included as an
  expert-assisted baseline. Row-wise best full-image results are in
  \textbf{bold}. Expert- and Model-branch results are shown in gray.}
  \label{tab:real_expert_evaluation}
  \setlength{\tabcolsep}{1.5pt}
  \renewcommand{\arraystretch}{0.78}
  \fontsize{7.2}{8.0}\selectfont

  \resizebox{\textwidth}{!}{%
  \begin{tabular}{llcccccccccc}
    \toprule
    \textbf{Dataset}
    & \textbf{Metric}
    & \textbf{CENet}
    & \textbf{MedSAM}
    & \textbf{MedSAM}
    & \textbf{PatchCorr}
    & \multicolumn{3}{c}{\textbf{Ours (1 expert)}}
    & \multicolumn{3}{c}{\textbf{Ours (3 experts)}} \\

    \multicolumn{2}{c}{}
    & \tiny{MICCAI 2025}
    & \tiny{NC 2024}
    & \tiny{Fine-tuned}
    & \tiny{MIDL 2024}
    & \textbf{System}
    & \textcolor{gray}{\textbf{Expert}}
    & \textcolor{gray}{\textbf{Model}}
    & \textbf{System}
    & \textcolor{gray}{\textbf{Expert}}
    & \textcolor{gray}{\textbf{Model}} \\
    \midrule

    \multirow{3}{*}{\shortstack{Chaksu}}
    & DSC
    & 97.20 $\pm$ 0.06
    & 94.97
    & 96.76 $\pm$ 0.08
    & 90.37 $\pm$ 0.07
    & \textbf{97.36 $\pm$ 0.05}
    & \textcolor{gray}{97.37 $\pm$ 0.04}
    & \textcolor{gray}{96.50 $\pm$ 0.10}
    & 97.17 $\pm$ 0.04
    & \textcolor{gray}{97.27 $\pm$ 0.05}
    & \textcolor{gray}{96.98 $\pm$ 0.15} \\

    & Jaccard
    & 94.64 $\pm$ 0.08
    & 90.68
    & 93.89 $\pm$ 0.13
    & 82.78 $\pm$ 0.04
    & \textbf{94.87 $\pm$ 0.01}
    & \textcolor{gray}{94.87 $\pm$ 0.03}
    & \textcolor{gray}{93.56 $\pm$ 0.08}
    & 94.60 $\pm$ 0.07
    & \textcolor{gray}{94.74 $\pm$ 0.11}
    & \textcolor{gray}{94.49 $\pm$ 0.09} \\

    & Sens.
    & 96.67 $\pm$ 0.15
    & 91.51
    & 96.69 $\pm$ 0.11
    & 84.11 $\pm$ 0.12
    & 96.30 $\pm$ 0.04
    & \textcolor{gray}{96.50 $\pm$ 0.04}
    & \textcolor{gray}{94.55 $\pm$ 0.05}
    & \textbf{96.74 $\pm$ 0.25}
    & \textcolor{gray}{95.72 $\pm$ 0.10}
    & \textcolor{gray}{96.84 $\pm$ 0.29} \\
    \midrule

    \multirow{3}{*}{\shortstack{Brain Growth\\(QUBIQ)}}
    & DSC
    & 86.76 $\pm$ 0.05
    & 66.98
    & 83.15 $\pm$ 0.50
    & 82.10 $\pm$ 0.47
    & 86.68 $\pm$ 0.93
    & \textcolor{gray}{92.32 $\pm$ 0.86}
    & \textcolor{gray}{84.87 $\pm$ 0.29}
    & \textbf{89.33 $\pm$ 1.34}
    & \textcolor{gray}{87.45 $\pm$ 2.06}
    & \textcolor{gray}{89.50 $\pm$ 1.79} \\

    & Jaccard
    & 79.10 $\pm$ 0.08
    & 50.57
    & 71.28 $\pm$ 0.67
    & 72.37 $\pm$ 0.75
    & 78.52 $\pm$ 1.37
    & \textcolor{gray}{86.11 $\pm$ 1.51}
    & \textcolor{gray}{73.96 $\pm$ 0.38}
    & \textbf{81.51 $\pm$ 1.89}
    & \textcolor{gray}{78.18 $\pm$ 3.34}
    & \textcolor{gray}{81.05 $\pm$ 2.45} \\

    & Sens.
    & 86.04 $\pm$ 0.23
    & 80.86
    & 85.03 $\pm$ 0.46
    & 84.38 $\pm$ 0.40
    & 86.49 $\pm$ 1.51
    & \textcolor{gray}{92.23 $\pm$ 1.03}
    & \textcolor{gray}{86.72 $\pm$ 1.12}
    & \textbf{88.20 $\pm$ 2.95}
    & \textcolor{gray}{86.14 $\pm$ 3.41}
    & \textcolor{gray}{90.46 $\pm$ 3.60} \\
    \midrule

    \multirow{3}{*}{\shortstack{Prostate\\(QUBIQ)}}
    & DSC
    & 94.23 $\pm$ 0.40
    & 93.39
    & 94.77 $\pm$ 0.20
    & 95.62 $\pm$ 0.04
    & \textbf{96.73 $\pm$ 0.28}
    & \textcolor{gray}{83.73 $\pm$ 2.07}
    & \textcolor{gray}{94.87 $\pm$ 0.53}
    & 95.71 $\pm$ 0.15
    & \textcolor{gray}{97.37 $\pm$ 0.39}
    & \textcolor{gray}{95.27 $\pm$ 1.44} \\

    & Jaccard
    & 89.36 $\pm$ 0.66
    & 89.24
    & 90.16 $\pm$ 0.37
    & 91.68 $\pm$ 0.09
    & \textbf{94.56 $\pm$ 0.49}
    & \textcolor{gray}{77.77 $\pm$ 3.81}
    & \textcolor{gray}{91.59 $\pm$ 0.91}
    & 92.17 $\pm$ 0.28
    & \textcolor{gray}{94.96 $\pm$ 0.74}
    & \textcolor{gray}{91.54 $\pm$ 2.52} \\

    & Sens.
    & \textbf{96.72 $\pm$ 0.46}
    & 92.16
    & 96.20 $\pm$ 0.32
    & 96.29 $\pm$ 0.17
    & 93.69 $\pm$ 1.32
    & \textcolor{gray}{87.35 $\pm$ 1.24}
    & \textcolor{gray}{95.57 $\pm$ 1.83}
    & 92.76 $\pm$ 0.73
    & \textcolor{gray}{97.80 $\pm$ 0.23}
    & \textcolor{gray}{93.12 $\pm$ 3.21} \\
    \midrule

    \multirow{3}{*}{\shortstack{Kidney\\(CURVAS)}}
    & DSC
    & 93.17 $\pm$ 0.43
    & 67.94
    & 90.70 $\pm$ 0.31
    & 91.32 $\pm$ 0.17
    & 94.48 $\pm$ 0.20
    & \textcolor{gray}{99.64 $\pm$ 0.26}
    & \textcolor{gray}{89.83 $\pm$ 0.21}
    & \textbf{95.80 $\pm$ 0.31}
    & \textcolor{gray}{99.37 $\pm$ 0.30}
    & \textcolor{gray}{89.58 $\pm$ 0.24} \\

    & Jaccard
    & 88.78 $\pm$ 0.60
    & 54.93
    & 84.41 $\pm$ 0.37
    & 86.75 $\pm$ 0.21
    & 90.00 $\pm$ 0.20
    & \textcolor{gray}{99.52 $\pm$ 0.29}
    & \textcolor{gray}{83.36 $\pm$ 0.19}
    & \textbf{91.98 $\pm$ 0.31}
    & \textcolor{gray}{99.12 $\pm$ 0.37}
    & \textcolor{gray}{83.09 $\pm$ 0.22} \\

    & Sens.
    & 92.45 $\pm$ 0.40
    & 88.90
    & 91.83 $\pm$ 0.66
    & 89.49 $\pm$ 0.18
    & 92.67 $\pm$ 0.23
    & \textcolor{gray}{99.64 $\pm$ 0.21}
    & \textcolor{gray}{90.09 $\pm$ 0.24}
    & \textbf{93.97 $\pm$ 0.54}
    & \textcolor{gray}{99.29 $\pm$ 0.32}
    & \textcolor{gray}{89.89 $\pm$ 0.51} \\
    \midrule

    \multirow{3}{*}{\shortstack{Liver\\(CURVAS)}}
    & DSC
    & 93.84 $\pm$ 0.16
    & 92.51
    & 93.14 $\pm$ 0.29
    & 92.39 $\pm$ 0.14
    & 93.55 $\pm$ 0.09
    & \textcolor{gray}{99.52 $\pm$ 0.07}
    & \textcolor{gray}{93.28 $\pm$ 0.20}
    & \textbf{95.80 $\pm$ 0.08}
    & \textcolor{gray}{99.47 $\pm$ 0.04}
    & \textcolor{gray}{93.13 $\pm$ 0.10} \\

    & Jaccard
    & 88.43 $\pm$ 0.28
    & 87.74
    & 88.39 $\pm$ 0.35
    & 87.24 $\pm$ 0.18
    & 89.93 $\pm$ 0.13
    & \textcolor{gray}{99.40 $\pm$ 0.09}
    & \textcolor{gray}{88.56 $\pm$ 0.23}
    & \textbf{91.94 $\pm$ 0.02}
    & \textcolor{gray}{99.15 $\pm$ 0.09}
    & \textcolor{gray}{88.39 $\pm$ 0.08} \\

    & Sens.
    & 93.65 $\pm$ 0.19
    & 92.68
    & 94.17 $\pm$ 0.18
    & 92.16 $\pm$ 0.09
    & \textbf{94.88 $\pm$ 0.24}
    & \textcolor{gray}{99.71 $\pm$ 0.08}
    & \textcolor{gray}{93.74 $\pm$ 0.36}
    & 94.04 $\pm$ 0.24
    & \textcolor{gray}{99.46 $\pm$ 0.11}
    & \textcolor{gray}{93.61 $\pm$ 0.30} \\

    \bottomrule
  \end{tabular}%
  }
\end{table*}

\section{Cross-center Validation on PROMISE12}
\label{app:cross_center}
To evaluate the robustness of $\nm$ under center shift, we conduct leave-one-center-out cross-center validation on the 50 labeled PROMISE12 training cases. PROMISE12 was originally collected from four acquisition centers with differences in scanner manufacturer, field strength, and imaging protocol. We denote the four site-specific subsets as Site A--D. In each fold, one site is held out for testing and the remaining sites are used for training. We compare MedSAM Zero-shot, supervised fine-tuned MedSAM, and $\nm$ with one or three expert branches.

As shown in Tab.~\ref{tab:cross_center_promise12_full}, the three-expert setting achieves the best DSC and Jaccard on all held-out sites. Compared with MedSAM Zero-shot and MedSAM-FT, $\nm$ consistently improves overlap-based performance, with the three-expert setting showing the largest gain in the case-weighted average. These results indicate that pixel-wise deferral can improve segmentation robustness under cross-center distribution shifts.

Tab.~\ref{tab:cross_center_workload} further reports the expert routing allocation. The three-expert setting defers only a small fraction of pixels to expert branches, ranging from 4.34\% to 6.25\% across held-out sites, with a case-weighted average deferral ratio of 5.73\%. This means that most pixels are still handled by the model branch. 

\begin{table*}[t!]
\centering
\caption{Leave-one-center-out cross-center validation on PROMISE12. Values are reported as percentages. MedSAM Zero-shot denotes direct inference using the original MedSAM checkpoint. MedSAM-FT denotes supervised fine-tuning on the training centers. The best result for each metric is highlighted in bold.}
\label{tab:cross_center_promise12_full}
\setlength{\tabcolsep}{2.0pt}
\renewcommand{\arraystretch}{0.8}
\resizebox{\textwidth}{!}{
\begin{tabular}{llcccccccccccc}
\toprule
\multirow{2}{*}{Held-out site} 
& \multirow{2}{*}{Cases}
& \multicolumn{3}{c}{MedSAM Zero-shot}
& \multicolumn{3}{c}{MedSAM-FT}
& \multicolumn{3}{c}{Ours-1E}
& \multicolumn{3}{c}{Ours-3E} \\
\cmidrule(lr){3-5}
\cmidrule(lr){6-8}
\cmidrule(lr){9-11}
\cmidrule(lr){12-14}
& & DSC & Jaccard & Sens.
& DSC & Jaccard & Sens.
& DSC & Jaccard & Sens.
& DSC & Jaccard & Sens. \\
\midrule
Site A 
& 15 
& 84.88 & 74.31 & \textbf{89.45}
& 86.10 & 76.16 & 81.23
& 87.16 & 77.24 & 82.46
& \textbf{91.48} & \textbf{84.95} & 85.30 \\

Site B 
& 15 
& 84.58 & 74.04 & 86.73
& 81.39 & 69.33 & 71.81
& 84.15 & 72.64 & 76.41
& \textbf{93.61} & \textbf{88.19} & \textbf{89.99} \\

Site C 
& 15 
& 83.46 & 72.23 & \textbf{88.00}
& 84.81 & 74.00 & 77.87
& 86.11 & 75.61 & 80.11
& \textbf{92.33} & \textbf{86.18} & 85.79 \\

Site D  
& 5 
& 80.89 & 68.57 & \textbf{89.24}
& 86.29 & 76.23 & 80.47
& 88.80 & 79.86 & 79.63
& \textbf{93.05} & \textbf{87.26} & 87.83 \\
\midrule
Center average
& -- 
& 83.45 & 72.29 & \textbf{88.35}
& 84.65 & 73.93 & 77.84
& 86.56 & 76.34 & 79.65
& \textbf{92.62} & \textbf{86.65} & 87.23 \\

Case-weighted average
& 50 
& 83.97 & 73.03 & \textbf{88.18}
& 84.32 & 73.47 & 77.32
& 86.11 & 75.63 & 79.66
& \textbf{92.53} & \textbf{86.52} & 87.11 \\
\bottomrule
\end{tabular}
}
\end{table*}

\begin{table}[t!]
\centering
\caption{Deferral ratio in leave-one-center-out cross-center validation on PROMISE12. Values are reported as percentages. Deferred denotes the percentage of pixels routed to expert branches.}
\label{tab:cross_center_workload}
\setlength{\tabcolsep}{3.0pt}
\renewcommand{\arraystretch}{0.92}
\fontsize{8pt}{8.5pt}\selectfont
\begin{tabular}{lccccccc}
\toprule
\multirow{2}{*}{Held-out site}
& \multicolumn{2}{c}{Ours-1E}
& \multicolumn{5}{c}{Ours-3E} \\
\cmidrule(lr){2-3}
\cmidrule(lr){4-8}
& Deferred & Model
& Deferred & E1 & E2 & E3 & Model \\
\midrule
Site A   & 2.53 & 97.47 & 6.06 & 0.62 & 4.31 & 1.13 & 93.94 \\
Site B        & 2.17 & 97.83 & 6.25  & 0.44 & 4.90 & 0.91 & 93.75 \\
Site C     & 1.95 & 98.05 & 5.34  & 0.52 & 3.84 & 0.99 & 94.66 \\
Site D   & 1.71 & 98.29 & 4.34  & 0.37 & 3.09 & 0.88 & 95.66 \\
\midrule
Center average     & 2.09 & 97.91 & 5.50  & 0.49 & 4.04 & 0.98 & 94.50 \\
Case-weighted average & 2.17 & 97.83 & 5.73 & 0.51 & 4.22 & 1.00 & 94.27 \\
\bottomrule
\end{tabular}
\end{table}
\section{Additional Qualitative Visualization}
\label{sec:appendix_additional_visualization}

\begin{figure}[!htbp]
\centering
\includegraphics[width=\textwidth,height=0.99\textheight,keepaspectratio]{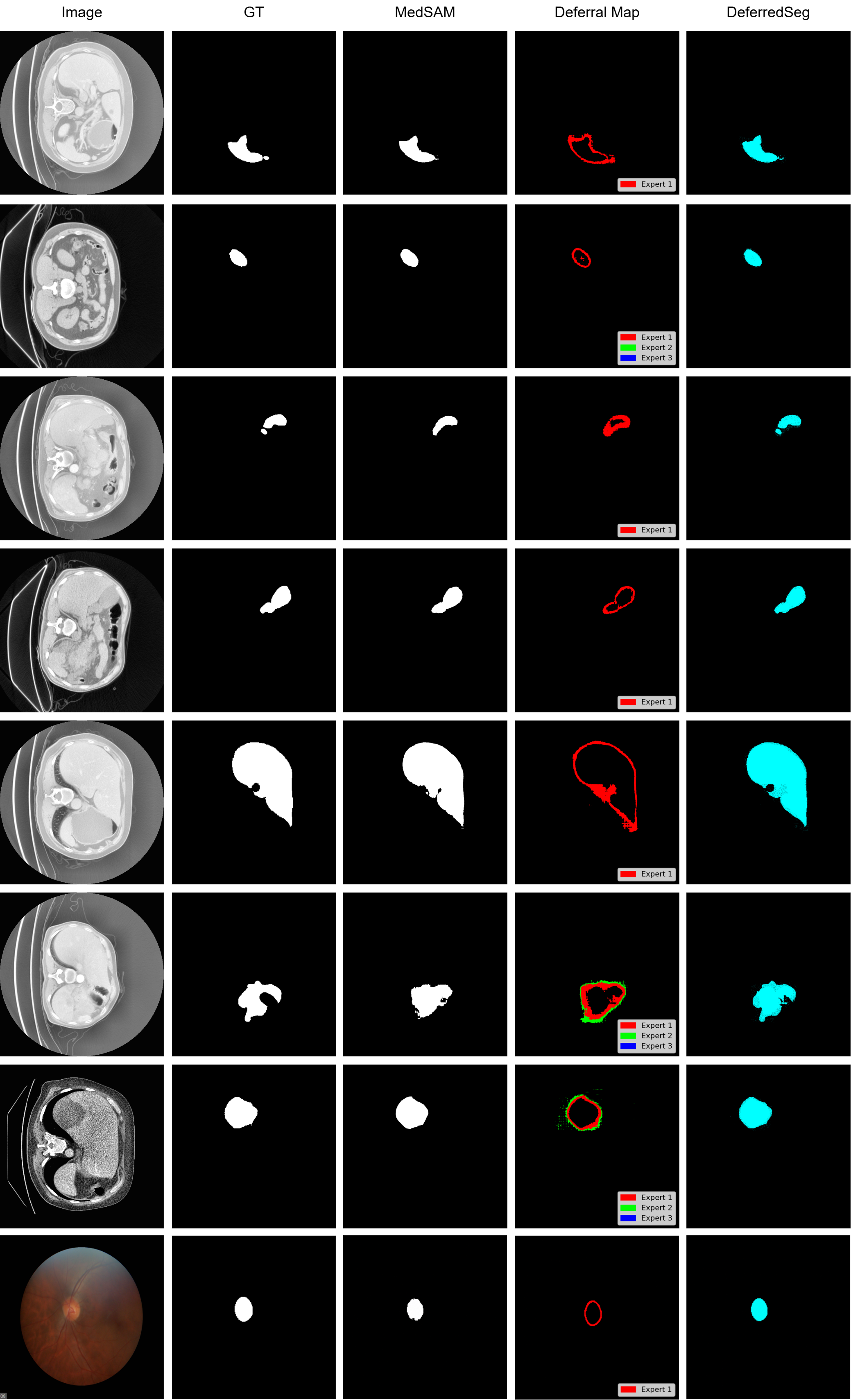}
\caption{Additional qualitative visualization results on different datasets, anatomical structures, and imaging modalities.}
\label{fig:appendix_visualization}
\end{figure}

To further demonstrate the generalizability of \(\nm\), we provide additional qualitative visualization results on multiple datasets and imaging modalities in Fig.~\ref{fig:appendix_visualization}. The supplementary examples cover multiple AMOS22 organs from abdominal CT/MRI scans, including spleen, right kidney, gallbladder, liver, and stomach, liver tumor CT cases from LiTS, and a retinal fundus example from Chaksu. Compared with the automated model branch, the System prediction of \(\nm\) better handles ambiguous boundaries and difficult foreground regions by learning to identify difficult pixels and defer them to more reliable expert branches. These examples complement Fig.~\ref{fig:seg_results_a} and provide broader qualitative evidence for the effectiveness and generalizability of the proposed Human--AI collaborative segmentation framework.

\section{Detailed Statistical Significance Analysis}
\label{app:statistical_significance}

Following the statistical testing protocol described in the main text, we
report the complete dataset-level results of the one-sided paired Wilcoxon
signed-rank tests on case-level DSC scores.

For the MedSAM-based experiments, the one- and three-expert configurations
of $\nm$ are separately compared with fine-tuned MedSAM (MedSAM-FT) and
PatchCorr on the same test cases. For the CENet-based experiments, CENet
with $\nm$ is compared with the corresponding fine-tuned CENet baseline
under the same backbone.

For each test case \(i\), W/T/L denotes whether the case-level DSC of
$\nm$ is higher than, equal to, or lower than that of baseline \(b\),
respectively:
\[
\mathrm{W}:
\mathrm{DSC}_{\nm,i}>\mathrm{DSC}_{b,i},
\qquad
\mathrm{T}:
\mathrm{DSC}_{\nm,i}=\mathrm{DSC}_{b,i},
\qquad
\mathrm{L}:
\mathrm{DSC}_{\nm,i}<\mathrm{DSC}_{b,i}.
\]
Thus, W/T/L reports the numbers of winning, tied, and losing test cases.
The null and alternative hypotheses are
\[
H_0:\Delta_b\leq 0,
\qquad
H_1:\Delta_b>0,
\]
where \(\Delta_b\) denotes the location shift of the paired DSC
differences relative to baseline \(b\). Therefore, \(H_1\) indicates that
$\nm$ achieves higher case-level DSC than the corresponding baseline.
Statistical significance is determined at \(p<0.05\), and n.s. denotes a
non-significant result.

\begin{table*}[!htbp]
\centering
\caption{Detailed statistical significance analysis of the MedSAM-based
one- and three-expert configurations using one-sided paired Wilcoxon
signed-rank tests on case-level DSC. Each configuration is compared with
fine-tuned MedSAM (MedSAM-FT) and PatchCorr on the same test cases.
W/T/L denotes the numbers of test cases in which $\nm$ obtains
higher/equal/lower case-level DSC than the corresponding baseline,
respectively. n.s. denotes \(p\geq0.05\).}
\label{tab:significance_medsam_full}

\setlength{\tabcolsep}{1.8pt}
\renewcommand{\arraystretch}{0.88}
\fontsize{7.4}{8.2}\selectfont

\resizebox{\textwidth}{!}{%
\begin{tabular}{lc|cc|cc|cc|cc}
\toprule
\multirow{3}{*}{Dataset}
& \multirow{3}{*}{Cases}
& \multicolumn{4}{c|}{One expert}
& \multicolumn{4}{c}{Three experts} \\
\cmidrule(lr){3-6}
\cmidrule(lr){7-10}

&
& \multicolumn{2}{c|}{vs. MedSAM-FT}
& \multicolumn{2}{c|}{vs. PatchCorr}
& \multicolumn{2}{c|}{vs. MedSAM-FT}
& \multicolumn{2}{c}{vs. PatchCorr} \\

&
& W/T/L & \(p\)
& W/T/L & \(p\)
& W/T/L & \(p\)
& W/T/L & \(p\) \\
\midrule

PROMISE12
& 30
& 29/0/1 & $<.001$
& 27/0/3 & $<.001$
& 29/0/1 & $<.001$
& 27/0/3 & $<.001$ \\

LiTS-Liver
& 28
& 27/0/1 & $<.001$
& 25/0/3 & $<.001$
& 15/0/13 & n.s.
& 21/0/7 & $<.01$ \\

LiTS-Tumor
& 27
& 26/0/1 & $<.001$
& 25/0/2 & $<.001$
& 25/0/2 & $<.001$
& 24/0/3 & $<.001$ \\

AMOS22-Spleen
& 120
& 96/0/24 & $<.001$
& 94/0/26 & $<.001$
& 99/0/21 & $<.001$
& 98/0/22 & $<.001$ \\

AMOS22-Right Kidney
& 118
& 104/0/14 & $<.001$
& 101/0/17 & $<.001$
& 103/0/15 & $<.001$
& 99/0/19 & $<.001$ \\

AMOS22-Left Kidney
& 120
& 108/0/12 & $<.001$
& 105/0/15 & $<.001$
& 106/0/14 & $<.001$
& 104/0/16 & $<.001$ \\

AMOS22-Gallbladder
& 112
& 103/0/9 & $<.001$
& 98/0/14 & $<.001$
& 106/0/6 & $<.001$
& 101/0/11 & $<.001$ \\

AMOS22-Esophagus
& 120
& 118/0/2 & $<.001$
& 112/0/8 & $<.001$
& 119/0/1 & $<.001$
& 115/0/5 & $<.001$ \\

AMOS22-Liver
& 120
& 63/0/57 & n.s.
& 112/0/8 & $<.001$
& 98/0/22 & $<.001$
& 116/0/4 & $<.001$ \\

AMOS22-Stomach
& 119
& 103/0/16 & $<.001$
& 101/0/18 & $<.001$
& 110/0/9 & $<.001$
& 108/0/11 & $<.001$ \\

AMOS22-Aorta
& 120
& 114/0/6 & $<.001$
& 111/0/9 & $<.001$
& 117/0/3 & $<.001$
& 115/0/5 & $<.001$ \\

AMOS22-Inferior Vena Cava
& 120
& 119/0/1 & $<.001$
& 115/0/5 & $<.001$
& 120/0/0 & $<.001$
& 116/0/4 & $<.001$ \\

AMOS22-Pancreas
& 120
& 117/0/3 & $<.001$
& 115/0/5 & $<.001$
& 119/0/1 & $<.001$
& 117/0/3 & $<.001$ \\

AMOS22-Duodenum
& 119
& 115/0/4 & $<.001$
& 113/0/6 & $<.001$
& 112/0/7 & $<.001$
& 110/0/9 & $<.001$ \\

AMOS22-Bladder
& 99
& 91/0/8 & $<.001$
& 89/0/10 & $<.001$
& 90/0/9 & $<.001$
& 87/0/12 & $<.001$ \\

AMOS22-Prostate/Uterus
& 97
& 94/0/3 & $<.001$
& 91/0/6 & $<.001$
& 93/0/4 & $<.001$
& 90/0/7 & $<.001$ \\

\bottomrule
\end{tabular}%
}
\end{table*}

As shown in Tab.~\ref{tab:significance_medsam_full}, the one- and
three-expert configurations each achieve statistically significant
improvements over MedSAM-FT in 15 of the 16 comparisons. The two
non-significant cases are AMOS22-Liver under the one-expert setting and
LiTS-Liver under the three-expert setting. Compared with PatchCorr, both
configurations achieve statistically significant improvements in all 16
dataset-level comparisons.

\begin{table}[!htbp]
\centering
\caption{Statistical significance analysis of the CENet-based
instantiation of $\nm$. One-sided paired Wilcoxon signed-rank tests are
performed on case-level DSC between CENet with $\nm$ and the corresponding
fine-tuned CENet baseline under the same backbone. W/T/L denotes the
numbers of test cases in which CENet with $\nm$ obtains
higher/equal/lower case-level DSC than CENet-FT, respectively. n.s.
denotes \(p\geq0.05\).}
\label{tab:significance_cenet_full}

\setlength{\tabcolsep}{5pt}
\renewcommand{\arraystretch}{0.92}

\begin{tabular}{lccc}
\toprule
Dataset & Cases & W/T/L & \(p\) \\
\midrule

PROMISE12
& 30
& 29/0/1
& $<.001$ \\

LiTS-Liver
& 28
& 20/0/8
& $<.05$ \\

LiTS-Tumor
& 27
& 22/0/5
& $<.01$ \\

AMOS22-Spleen
& 120
& 98/0/22
& $<.001$ \\

Right Kidney
& 118
& 101/0/17
& $<.001$ \\

Chaksu
& 273
& 206/0/67
& $<.05$ \\

\bottomrule
\end{tabular}
\end{table}

As shown in Tab.~\ref{tab:significance_cenet_full}, adding $\nm$ to the
CENet backbone produces statistically significant improvements over
CENet-FT on all six evaluated segmentation tasks.


\section{Detailed Ablation Results}
\label{sec:appendix_ablation}

As a complement to Tab.~\ref{tab:ablation} in the main text, Tab.~\ref{tab:ablation_all} reports the complete ablation results, including the System, Expert, and Model branches under different loss configurations. These detailed results further verify that the proposed components are mutually complementary and indispensable.

\begin{table}[!htbp]
  \centering
  \scriptsize
  \caption{Analysis of each loss component through ablation. These components are mutually complementary and indispensable.}
  \label{tab:ablation_all}
  \resizebox{\linewidth}{!}{%
    \begin{tabular}{ll|ccc|ccc|ccc}
      \toprule
        \textbf{J} & \textbf{Loss} 
          & \multicolumn{3}{c|}{\textbf{DSC}} 
          & \multicolumn{3}{c|}{\textbf{Jaccard}} 
          & \multicolumn{3}{c}{\textbf{Sens.}} \\
        & & System & Expert & Model 
          & System & Expert & Model 
          & System & Expert & Model \\
        \midrule
        \multirow{2}{*}{1}
          & DC 
            & 96.86$\pm$0.92 & 95.67$\pm$1.48 & 94.55$\pm$0.89
            & 94.00$\pm$1.73 & 91.78$\pm$2.72 & 93.67$\pm$1.01
            & 96.80$\pm$1.89 & 95.32$\pm$2.90 & 96.49$\pm$1.71 \\
          & + SC
            & \textbf{97.17$\pm$0.63} & \textbf{96.09$\pm$0.88} & \textbf{96.08$\pm$0.52}
            & \textbf{94.57$\pm$1.17} & \textbf{92.55$\pm$1.62} & \textbf{95.14$\pm$0.54}
            & \textbf{97.29$\pm$1.24} & \textbf{96.01$\pm$1.76} & \textbf{97.41$\pm$1.32} \\
        \midrule
        \multirow{4}{*}{3}
          & DC
            & 95.39$\pm$1.64 & 93.49$\pm$2.63 & 94.35$\pm$1.41 
            & 91.51$\pm$2.95 & 88.23$\pm$4.57 & 93.58$\pm$1.46 
            & 97.13$\pm$2.23 & 95.76$\pm$3.51 & \textbf{98.42$\pm$1.84} \\
          & + SC
            & 95.79$\pm$1.71 & 93.98$\pm$2.58 & 94.16$\pm$0.76 
            & 92.14$\pm$3.07 & 88.98$\pm$4.51 & 93.38$\pm$0.70 
            & 97.18$\pm$2.31 & 95.79$\pm$3.55 & 97.39$\pm$3.01 \\
          & + LB
            & \textbf{96.59$\pm$1.46} & 95.06$\pm$2.49 & 95.72$\pm$0.84 
            & 93.52$\pm$2.69 & 90.78$\pm$4.45 & 94.70$\pm$0.89 
            & 97.16$\pm$2.19 & 95.78$\pm$3.52 & 97.94$\pm$1.55 \\
          & + SC, LB
            & 96.58$\pm$1.04 & \textbf{95.08$\pm$1.78} & \textbf{96.19$\pm$0.55}
            & \textbf{93.56$\pm$1.87} & \textbf{90.85$\pm$3.08} & \textbf{95.25$\pm$0.56}
            & \textbf{98.34$\pm$0.47} & \textbf{97.61$\pm$0.68} & 98.39$\pm$1.03 \\
        \bottomrule
    \end{tabular}
  }
\end{table}
\section{Full Results of Coherence-Balance grid search}\label{app:coherence_balance}

Table~\ref{tab:comparison_lambdas} reports the complete numerical results for all branches (System / Expert / Model) and metrics (Jaccard / DSC / Sensitivity) across all $\lambda_1,\lambda_2$ combinations.

\begin{table}[!htbp]
\centering
\caption{Effect of varying coherence ($\lambda_1$) and balance ($\lambda_2$) weights on System / Expert / Model performance (\%).}
\label{tab:comparison_lambdas}
\resizebox{\linewidth}{!}{%
\begin{tabular}{l|cc|ccc|ccc|ccc}
\toprule
Group & $\lambda_1$ & $\lambda_2$
  & \multicolumn{3}{c|}{\textbf{System}}
  & \multicolumn{3}{c|}{\textbf{Expert}}
  & \multicolumn{3}{c}{\textbf{Model}} \\
  
& & & Jaccard & DSC & Sens.
    & Jaccard & DSC & Sens.
    & Jaccard & DSC & Sens. \\
\midrule
A & 0.10 & 0.10 & 91.26 & 95.35 & 94.74 & 87.45 & 93.21 & 91.99 & 92.71 & 93.32 & 98.55 \\
B & 0.10 & 1.00 & 91.12 & 95.25 & 94.73 & 87.51 & 93.25 & 92.00 & 93.80 & 94.43 & \textbf{99.90} \\
C & 0.10 & 5.00  & 91.04 & 95.11 & 94.76 & 87.42 & 93.09 & 91.99 & 93.66 & 94.47 & 95.20 \\
D & 0.10 & 10.00 & 91.36 & 95.43 & 94.67 & 87.89 & 93.50 & 92.02 & 93.12 & 93.95 & 99.23 \\
E & 1.00 & 0.10& 91.37 & 95.44 & 94.66 & 87.96 & 93.54 & 91.99 & 93.02 & 93.84 & 99.39 \\
F & 1.00 & 1.00 & 90.74   & 95.02 & 94.66 & 87.08 & 92.97 & 91.97 & 93.85  & 94.59  & 99.45\\
G & 1.00 & 5.00 & \textbf{91.90} & \textbf{95.63} & \textbf{96.13} & \textbf{88.48} & \textbf{93.73} & \textbf{93.98} & 93.20 & 94.32 & 94.86 \\
H & 1.00 & 10.00 & 91.03 & 95.21 & 94.70 & 87.27 & 93.09 & 92.01 & 93.47 & 94.21 & 98.91 \\
I & 5.00 & 0.10& 90.28 & 94.61 & 94.77 & 86.43 & 92.41 & 92.00 & 93.70 & 94.44 & 99.59 \\
J & 5.00 & 1.00 & 91.63 & 95.58 & 94.83 & 87.90 & 93.50 & 92.01 & 92.57 & 93.30 & 99.09 \\
K & 5.00 & 5.00& 91.12 & 95.22 & 94.66 & 87.71 & 93.33 & 91.99 & 93.71 & 94.51 & 95.02 \\
L & 5.00 & 10.00 & 91.64 & 95.59 & 94.60 & 88.22 & 93.69 & 92.01 & 91.76 & 92.42 & 99.24 \\
M & 10.00 & 0.10 & 91.30 & 95.36 & 94.64 & 87.88 & 93.45 & 92.01 & 92.14 & 92.94 & 98.18 \\
N & 10.00 & 1.00 & 91.13 & 95.28 & 94.62 & 87.65 & 93.35 & 91.98 & 92.20 & 93.19 & 98.39 \\
O & 10.00 & 5.00 & 91.22 & 95.34 & 94.68 & 87.64 & 93.32 & 92.01 & \textbf{94.06} & \textbf{94.88} & 99.49 \\
P & 10.00 & 10.00 & 90.53 & 94.87 & 94.68 & 86.91 & 92.84 & 92.00 & 90.66 & 91.32 & 99.55 \\
\bottomrule
\end{tabular}
}
\end{table}

\begin{figure}[t]
  \centering
  \includegraphics[width=\linewidth]{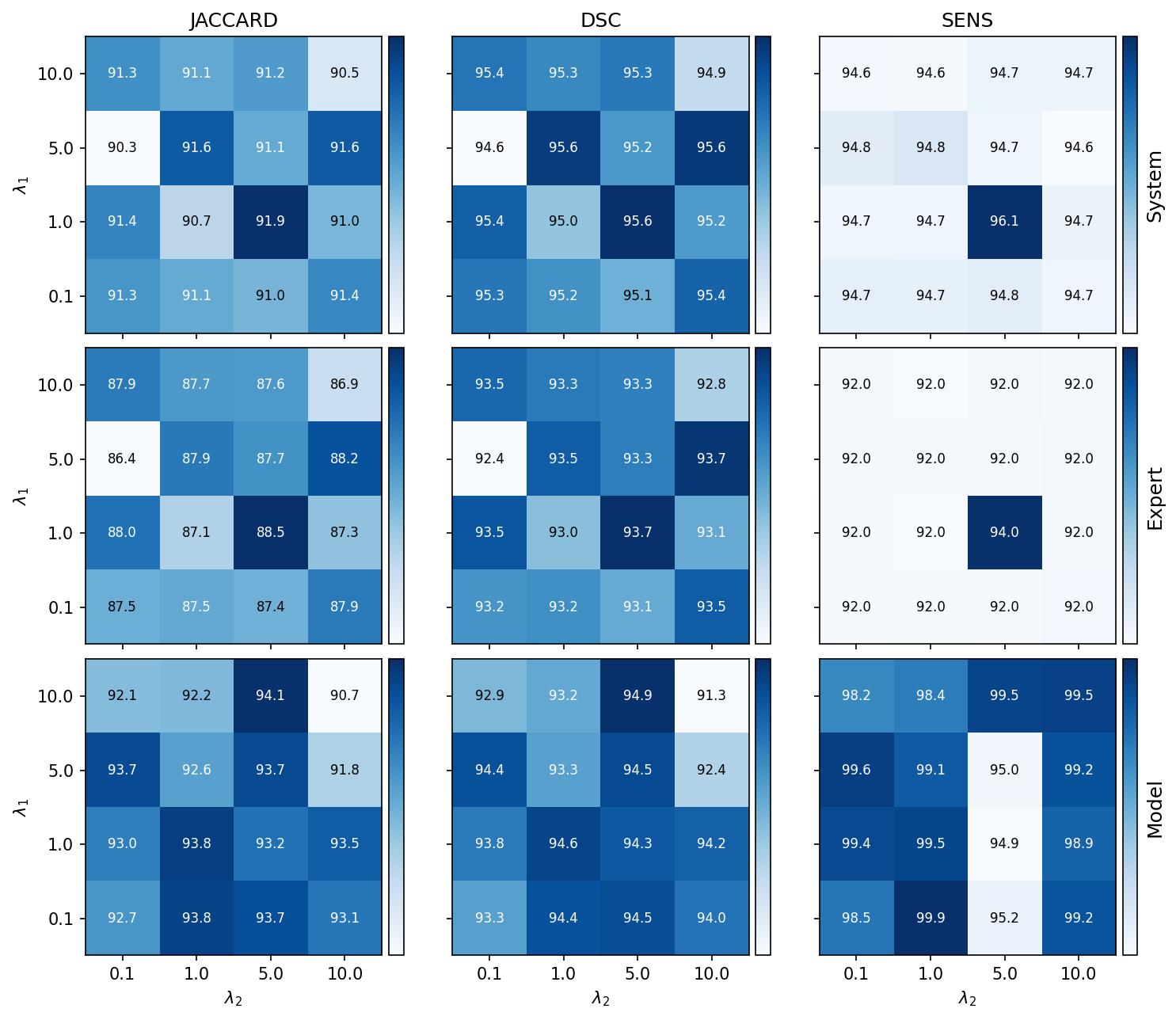}
  \caption{Complete set of heatmaps for all branches and metrics.}
  \label{fig:lambda_heatmap_9}
\end{figure}

\textbf{Full Heatmaps:}
For completeness, Fig.~\ref{fig:lambda_heatmap_9} shows the \emph{nine} heatmaps spanning all branches (System / Expert / Model) and metrics (Jaccard / DSC / Sensitivity). 

\subsection{Additional Hyperparameter Analysis on More Datasets}
\label{app:hyperparameter_more_datasets}

To further examine the robustness of the hyperparameter choice, we conduct an additional sensitivity analysis on LiTS-Liver under the three-expert setting by varying the coherence and balance weights $(\lambda_1,\lambda_2)$. As shown in Tab.~\ref{tab:hyperparameter_lits_liver}, the default setting $(\lambda_1,\lambda_2)=(1,5)$, selected from the PROMISE12 grid search, achieves the best overlap-based performance among the tested configurations, with the highest DSC and Jaccard.

\begin{table}[!htbp]
\centering
\caption{Effect of varying coherence ($\lambda_1$) and balance ($\lambda_2$) weights on System performance on LiTS-Liver (\%).}
\label{tab:hyperparameter_lits_liver}
\begin{tabular}{l|cc|ccc}
\toprule
Group & $\lambda_1$ & $\lambda_2$
  & \multicolumn{3}{c}{\textbf{System}} \\
& & & DSC & Jaccard & Sens. \\
\midrule
A & 0.10 & 0.10 & 94.15 & 88.95 & 92.97 \\
B & 0.10 & 5.00 & 95.82 & 91.98 & 94.10 \\
C & 1.00 & 1.00 & 95.95 & 92.22 & 94.77 \\
D & 1.00 & 5.00 & \textbf{96.15} & \textbf{93.70} & \textbf{95.47} \\
E & 1.00 & 10.00 & 96.01 & 92.33 & 94.88 \\
F & 5.00 & 1.00 & 94.66 & 89.86 & 93.48 \\
G & 5.00 & 5.00 & 93.86 & 88.43 & 92.68 \\
H & 10.00 & 10.00 & 91.10 & 83.65 & 89.92 \\
\bottomrule
\end{tabular}
\end{table}

\section{Expert Scalability Analysis}\label{app:scalability_table}
For the scalability experiments, each expert set is drawn from a fixed pool of five synthetic experts (E1-E5), respectively (see Tab.~\ref{tab:scalability_expert_settings}), while keeping the overall FG/BG/BD accuracy approximately constant across groups. 
Specifically:
\begin{itemize}
  \item 1 expert uses \textbf{E1}.
  \item 2 experts use \textbf{E2, E3}.
  \item 3 experts use \textbf{E1, E2, E3}.
  \item 4 experts use \textbf{E2, E3, E4, E5}.
  \item 5 experts use \textbf{E1, E2, E3, E4, E5}.
\end{itemize}

The complete System/Expert/Model scores are provided in Tab.~\ref{tab:num_experts_horizontal}

\begin{table}[!htbp]
  \centering
  \caption{Settings of scalability-specific synthetic expert pool.}
  \label{tab:scalability_expert_settings}
  \begin{tabular}{c|ccc}
    \toprule
    \textbf{Expert} & \textbf{FG} & \textbf{BG} & \textbf{BD} \\
    \midrule
    E1 & 92 & 98 & 97 \\
    E2 & 91 & 99 & 96 \\
    E3 & 93 & 97 & 98 \\
    E4 & 90 & 97 & 99 \\
    E5 & 94 & 99 & 99 \\
    \bottomrule
  \end{tabular}
\end{table}

\begin{table}[!htbp]
  \centering
  \caption{Performance on PROMISE12 under different numbers of experts (\%). 
Single expert yields peak overlap (DSC/Jaccard), whereas multiple experts raise Sensitivity. 
(mean $\pm$ std; row-wise best in \textbf{bold}, second best in \textit{italics})}

  \label{tab:num_experts_horizontal}
  \setlength{\tabcolsep}{6pt}
  \begin{adjustbox}{max width=\linewidth}
  \begin{tabular}{c|c|ccccc}
    \toprule
    \textbf{Metric} & \textbf{Predictor} & \textbf{J = 1} & \textbf{J = 2} & \textbf{J = 3} & \textbf{J = 4} & \textbf{J = 5} \\
    \midrule
    \multirow{3}{*}{DSC} 
      & System & \textbf{97.06 $\pm$ 0.05} & 94.48 $\pm$ 0.49 & \textit{96.00 $\pm$ 0.14} & 94.43 $\pm$ 0.52 & 95.45 $\pm$ 0.31 \\
      & Expert & \textbf{97.88 $\pm$ 0.01} & 92.08 $\pm$ 0.72 & \textit{95.09 $\pm$ 0.19} & 93.05 $\pm$ 0.57 & 93.48 $\pm$ 0.27\\
      & Model  & 82.80 $\pm$ 2.75 & \textbf{92.14 $\pm$ 2.60} & 87.17 $\pm$ 1.99 & 82.44 $\pm$ 0.61 & \textit{91.45 $\pm$ 1.84}\\
    \midrule
    \multirow{3}{*}{Jaccard} 
      & System & \textbf{94.38 $\pm$ 0.08} & 89.65 $\pm$ 0.83 & \textit{92.46 $\pm$ 0.24} & 89.63 $\pm$ 0.89 & 91.44 $\pm$  0.64\\
      & Expert & \textbf{95.85 $\pm$ 0.01} & 85.42 $\pm$ 1.20 & \textit{90.80 $\pm$ 0.32} & 87.16 $\pm$ 0.96 & 87.94 $\pm$  0.94\\
      & Model  & 79.56 $\pm$ 2.68 & \textbf{91.05 $\pm$ 2.30} & 86.80 $\pm$ 1.91 & 82.24 $\pm$ 0.61 & \textit{90.71 $\pm$  1.62} \\
    \midrule
    \multirow{3}{*}{Sens.} 
      & System & 95.33 $\pm$ 0.09 & 94.08 $\pm$ 0.13 & \textit{96.42 $\pm$ 0.04} & 94.76 $\pm$ 0.06 & \textbf{96.64 $\pm$ 0.11}\\
      & Expert & \textbf{97.01 $\pm$ 0.00} & 91.07 $\pm$ 0.01 & \textit{95.57 $\pm$ 0.04} & 93.29 $\pm$ 0.08 & 95.00 $\pm$  0.06\\
      & Model  & 83.55 $\pm$ 1.37 & \textit{99.21 $\pm$ 0.43} & 99.05 $\pm$ 0.64 & \textbf{99.75 $\pm$ 0.25} & 91.95 $\pm$  0.19\\
    \bottomrule
  \end{tabular}
  \end{adjustbox}
\end{table}

\section{Complementary Expert Settings and Detailed Results}
\label{sec:appendix_complementary_experts}
To analyze the impact of complementary expert configurations, we define seven synthetic experts with distinct accuracy profiles on FG, BG, and BD regions (Tab.~\ref{tab:complementary_experts}). By selecting different subsets of these experts, we emulate varying levels of complementarity and specialization.

\begin{table}[!htbp]
  \centering
  \caption{Accuracy profiles of complementary experts.}
  \label{tab:complementary_experts}
  \begin{adjustbox}{max width=\linewidth}
  \begin{tabular}{l|ccc}
    \toprule
    \textbf{Expert Type} & \textbf{FG} & \textbf{BG} & \textbf{BD} \\
    \midrule
    (E1) FG strong          & 0.97 & 0.90 & 0.98 \\
    (E2) BG strong          & 0.88 & 0.99 & 0.96 \\
    (E3) BD strong          & 0.88 & 0.90 & 1.00 \\
    (E4) FG+BG strong       & 0.97 & 0.99 & 0.98 \\
    (E5) FG+BD strong       & 0.97 & 0.90 & 1.00 \\
    (E6) BG+BD strong       & 0.88 & 0.99 & 1.00 \\
    (E7) FG+BG+BD strong    & 0.97 & 0.99 & 1.00 \\
    \bottomrule
  \end{tabular}
  \end{adjustbox}
\end{table}

Table~\ref{tab:complementary_num_results} presents the detailed performance of System / Expert / Model branches as the number of complementary experts increases.  
A single complementary expert (J=1) already achieves high System performance. 
Adding three or more complementary experts (J=3-5) further improves System Jaccard and DSC, with the strongest results obtained at J=5 (Tab.~\ref{tab:complementary_num_results}).

\begin{table}[!htbp]
  \centering
  \caption{System / Expert / Model performance with increasing number of complementary experts (mean $\pm$ std, \%).}
  \label{tab:complementary_num_results}
  \setlength{\tabcolsep}{2.0pt}
\renewcommand{\arraystretch}{0.8}
  \small
  \resizebox{\textwidth}{!}{%
    \begin{tabular}{c|l|ccc|ccc|ccc}
      \toprule
      J & Experts & \multicolumn{3}{c|}{System} & \multicolumn{3}{c|}{Expert} & \multicolumn{3}{c}{Model} \\
      \midrule
       & & Jaccard & DSC & Sens. & Jaccard & DSC & Sens. & Jaccard & DSC & Sens. \\
      \midrule
      1 & E1
        & 91.04 $\pm$ 0.16 & 95.24 $\pm$ 0.10 & 95.54 $\pm$ 0.06
        & 91.00 $\pm$ 0.19 & 95.21 $\pm$ 0.10 & 97.01 $\pm$ 0.02
        & 80.81 $\pm$ 2.70 & 83.94 $\pm$ 2.84 & 85.12 $\pm$ 1.20 \\
      2 & E1, E6
        & 89.03 $\pm$ 0.72 & 94.10 $\pm$ 0.42 & 93.58 $\pm$ 0.81
        & 91.91 $\pm$ 0.09 & 95.73 $\pm$ 0.05 & \textbf{97.04 $\pm$ 0.00}
        & 84.07 $\pm$ 5.16 & \textbf{89.49 $\pm$ 5.01} & 91.85 $\pm$ 2.05 \\
      3 & E1, E2, E3
        & 92.45 $\pm$ 0.21 & 95.99 $\pm$ 0.13 & 96.38 $\pm$ 0.05
        & 90.73 $\pm$ 0.26 & 95.05 $\pm$ 0.15 & 95.51 $\pm$ 0.01
        & \textbf{85.21 $\pm$ 2.68} & 85.59 $\pm$ 2.66 & 98.76 $\pm$ 0.97 \\
      4 & E1, E2, E3, E4
        & 94.06 $\pm$ 0.46 &  96.83 $\pm$ 0.18 & 97.43 $\pm$ 0.04
        & 92.02 $\pm$ 0.18 &  95.50 $\pm$ 0.12 & 96.94 $\pm$ 0.01
        & 74.45 $\pm$ 3.00 &  74.62 $\pm$ 2.98 & \textbf{99.50 $\pm$ 0.44} \\
      5 & E1, E2, E3, E4, E5
        & \textbf{94.40 $\pm$ 0.45} & \textbf{96.88 $\pm$ 0.26} & \textbf{97.46 $\pm$ 0.07}
        & \textbf{93.13 $\pm$ 0.36} & \textbf{96.39 $\pm$ 0.21} & 96.92 $\pm$ 1.42
        & 78.31 $\pm$ 1.58 & 85.84 $\pm$ 3.13 & 98.48 $\pm$ 0.03 \\
      \bottomrule
    \end{tabular}%
  }
\end{table}


\section{Additional Computational Cost Analysis}
\label{sec:computational_overhead}

We further evaluate the computational cost of $\nm$ with one and three
experts against the base MedSAM under the same hardware, precision,
batch-size, and input-resolution settings. As shown in
Tab.~\ref{tab:computational_cost}, the one-expert configuration increases
the parameter count and FLOPs by only 0.63\% and 0.24\%, respectively,
while increasing the average inference time by 4.33\%. The three-expert
configuration increases the parameter count and FLOPs by 0.94\% and
0.26\%, respectively, with a 5.94\% increase in inference time. These
results show that $\nm$ introduces only a small increase in model
complexity and computational cost compared with the base MedSAM, while
providing the additional deferral capability.

\begin{table}[htbp]
\centering
\caption{Additional computational-cost comparison with the base MedSAM.
FLOPs and inference time are measured with an input resolution of
$1024\times1024$, a batch size of 1, and FP32 precision on the same GPU.
Inference time is reported as the mean and standard deviation over five
repeated runs.}
\label{tab:computational_cost}
\setlength{\tabcolsep}{5pt}
\begin{tabular}{lccc}
\toprule
Method & Params (M) & FLOPs (G) & Time (ms/image) \\
\midrule
MedSAM
& 90.7372
& 69.2878
& 16.1161 $\pm$ 0.0223 \\

$\nm$ (1 expert)
& 91.3125
& 69.4532
& 16.8139 $\pm$ 0.0157 \\

$\nm$ (3 experts)
& 91.5926
& 69.4676
& 17.0741 $\pm$ 0.0086 \\
\bottomrule
\end{tabular}
\end{table}
\end{document}